\documentclass[11pt,a4paper]{article}

\usepackage[T1]{fontenc}
\usepackage[english]{babel}

\usepackage[margin=3cm]{geometry}
\usepackage{amsmath,amssymb}
\usepackage{graphicx}
\usepackage{xcolor}
\usepackage{booktabs}
\usepackage{enumitem}
\usepackage{array}
\usepackage{tabularx}
\usepackage{float}
\usepackage{changepage}
\usepackage{microtype}
\usepackage[most]{tcolorbox}
\usepackage[hidelinks,breaklinks=true]{hyperref}
\usepackage{xurl}

\usepackage[numbers,sort&compress]{natbib}

\renewenvironment{abstract}{%
  \small
  \begin{center}\bfseries\abstractname\end{center}%
  \vspace{0.6em}%
  \begin{adjustwidth}{2em}{2em}%
}{%
  \end{adjustwidth}\vspace{1.2em}%
}

\definecolor{accent}{RGB}{9,20,120}

\newtcolorbox{axe}[1][]{%
  enhanced, breakable,
  colback=accent!3, colframe=accent!3,
  borderline west={2.2pt}{0pt}{accent},
  boxrule=0pt, arc=0pt, outer arc=0pt,
  left=11pt, right=11pt, top=8pt, bottom=8pt,
  fontupper=\small,
  #1
}

\newcommand{\KL}{\operatorname{KL}}

\title{\bfseries One Symptom, Three Levers: A Critical Review of On-Policy Self-Distillation}
\author{%
  Justin Robert\thanks{Corresponding author: \texttt{robert.just@yahoo.com}.} \qquad Raheel Qader\\[8pt]
  \normalsize OVHai LLM
}
\date{}

\begin{document}
\maketitle
\vspace{-2.2em}

\begin{abstract}
\noindent
On-policy distillation trains a language model on its own generations while a teacher scores them token by token. It combines the dense supervision of imitation learning with the on-policy sampling of reinforcement learning. But it requires a second, larger model to act as teacher. On-Policy Self-Distillation (OPSD) removes that cost. The teacher is the model itself, conditioned on privileged information the student will not have at test time, such as a reference solution, a plan, or environment feedback. The teacher is no stronger than the student, only better informed. Early results were promising, with accuracy comparable to reinforcement learning at a fraction of the generated tokens. But the same asymmetry that produces the signal also biases it. One failure mode now dominates the field: collapse, the progressive narrowing of the set of reasoning paths the model can produce. Collapse is not specific to OPSD, though privileged information aggravates it. This review treats collapse as a symptom governed by three levers: (i) where the signal is applied, that is, how tokens are weighted; (ii) what the teacher is shown, that is, the nature of the privileged information; and (iii) when the signal changes, that is, the teacher's dynamics and the decay of guidance. We restrict our scope to mathematical reasoning, where the method originated and where its failure modes are best documented. We report no new experiments. The contribution is structural: a shared vocabulary for phenomena named differently across papers, and a clear line between what is settled and what is still disputed.
\end{abstract}


\section*{Introduction}

Since the release of DeepSeek-R1 \citep{deepseekr1}, reinforcement learning with verifiable rewards (RLVR) has become the dominant route to giving large language models reasoning ability. The model generates several rollouts, a reward is issued according to the correctness of the final answer, and a GRPO-style algorithm \citep{shao2024grpo} updates the policy. The approach has produced striking progress, but structural limits remain. The reward is sparse: a single signal at the end of the trajectory must account for hundreds of tokens, which makes credit assignment difficult. It is expensive, since many long rollouts must be sampled for each problem. And it tends to concentrate probability mass on reasoning the base model already produces, rather than discovering new reasoning.

\medskip
On-policy distillation (OPD) was developed to restore a dense signal without giving up on-policy sampling \citep{agarwal2024gkd, lu2025blog}. A teacher scores the student's rollouts token by token, which resolves credit assignment on the
distribution the student actually visits. The price is a second model, larger than the student, that must run alongside it throughout training.

\medskip
On-Policy Self-Distillation (OPSD) \citep{zhao2026opsd} and Self-Distillation
Policy Optimization (SDPO) \citep{hubotter2026sdpo}, proposed independently within days of each other, remove that dependency. The teacher is the model itself, given privileged information that the student does not receive: a reference solution for OPSD, feedback from the environment for SDPO. The teacher therefore holds more information than the student, which lets it produce a dense token-level signal without any larger model.

\bigskip
Combining the signal density of distillation with the autonomy of reinforcement learning (RL) makes it possible, in principle, to train small reasoning models on a reduced budget and without depending on a larger model. That promise comes with unresolved tensions. The dense signal can collapse the model's diversity and entropy. It can also degrade capabilities acquired earlier, and lead the student to rely on information it will not have at test time.

\paragraph{Scope and method.} This paper does not aim for exhaustiveness. The field opened by OPSD now comprises more than two hundred works. Its two largest branches are multimodal learning and tool-using agents. Both obey the same tensions, with domain-specific instantiations that we do not cover. We focus on mathematical reasoning, where the method was introduced and where its failure modes are best documented. Readers seeking an exhaustive map of on-policy distillation, external teachers included, should turn to the surveys of \citet{song2026survey} and \citet{zhang2026formula}. A brief overview of OPSD exists \citep{cui2026overview}, but it organizes the field by families of methods and addresses neither failure modes nor open questions. This paper covers work available up to August 2026.

\bigskip \textbf{Since the founding paper, how has the field learned to control the dense signal that OPSD produces?} Part 1 lays the foundations: where the method comes from, how it works, and what its founding paper leaves open. Part 2 takes the symptom and the three levers one at a time, separating for each what is settled from what is still disputed. Part 3 brings them together: where to apply the signal, what to show the teacher, and when to let the guidance change.


\bigskip
\section{OPSD: Where It Comes From, How It Works, What It Is For}
\label{sec:partie1}

\subsection{Genealogy: From RL and Distillation to OPSD}
\label{sec:background}

OPSD is the endpoint of a sequence in which each post-training method corrects
the shortcoming of the previous one. A single question runs through them:
\textbf{how can a model be given a dense, cheap learning signal without
depending on a larger model?}

\bigskip
The starting point is \textbf{supervised fine-tuning} (SFT), in which the model
imitates reasoning traces token by token. The signal is \textit{dense}, but it
applies to sequences the model does not produce itself (\textit{off-policy}).
Because it is trained to continue correct prefixes, the model drifts at
inference time as soon as it leaves them. This is \emph{exposure bias}.

\bigskip
\textbf{Reinforcement learning} removes that obstacle by optimizing the model's
own generations (\textit{on-policy}). It was initially based on human feedback
(RLHF): annotators ranked rollouts by preference, and those rankings trained a
reward model that imitated human judgment. The procedure was slow and
expensive. RLVR replaced it by restricting training to problems whose answer
can be checked automatically, such as mathematics and code, so that rollouts
are ranked without human intervention.\\
The model now learns from a rollout it generated itself, but the signal is no
longer dense. It receives a single reward, at the end of the trajectory, for
hundreds of tokens. Credit assignment becomes difficult again, and sampling
long rollouts is expensive.

\begin{figure}[H]
    \centering
    \includegraphics[width=0.85\linewidth]{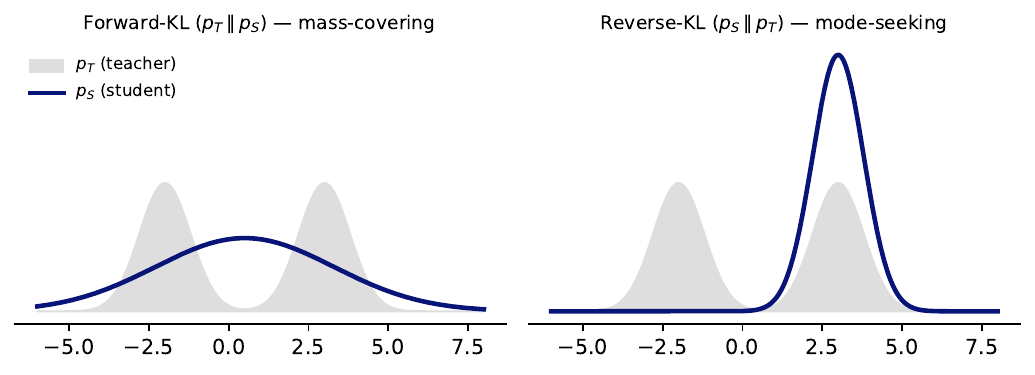}
    \caption{Forward KL versus reverse KL.}
    \label{fig:fr-kl}
\end{figure}

\bigskip
\textbf{On-policy distillation} restores that density. A teacher model scores
the rollouts the student generates. Generalized Knowledge Distillation (GKD)
\citep{agarwal2024gkd} formalizes the scheme: for each token generated by the
student, the teacher returns its next-token probability distribution over the
whole vocabulary. The two distributions are compared, and the gap between them
is reduced over successive steps, which brings the student towards the
teacher's capabilities.\\
One choice matters throughout the paper. Two distributions can be brought
together in two opposite directions:
\begin{itemize}
    \item the forward KL pushes the student to cover all of the teacher's modes;
    \item the reverse KL pushes the student to settle on a single one of them
    (Figure~\ref{fig:fr-kl}, \S\ref{sec:A1}).
\end{itemize}
The latter, popularized for LLMs by MiniLLM \citep{gu2024minillm}, reduces
exposure bias but can impoverish diversity (\S\ref{sec:A1}). One dependency
still remains: the teacher is a larger model, and therefore expensive to run.

\bigskip
\textbf{On-Policy Self-Distillation} (OPSD) removes it. Rather than calling on
a larger model, it uses the model itself, given privileged information
$y^\star$: a reference solution, a hint, or environment feedback. The principle
has a long history. It instantiates the theory of learning using privileged
information introduced by \citet{vapnik2009lupi} in 2009, and the literature
has repeatedly shown that a model can be improved from a signal derived from
itself \citep{furlanello2018ban, mobahi2020self, zelikman2022star}.

\begin{figure}[htbp]
    \centering
    \includegraphics[width=0.65\textwidth]{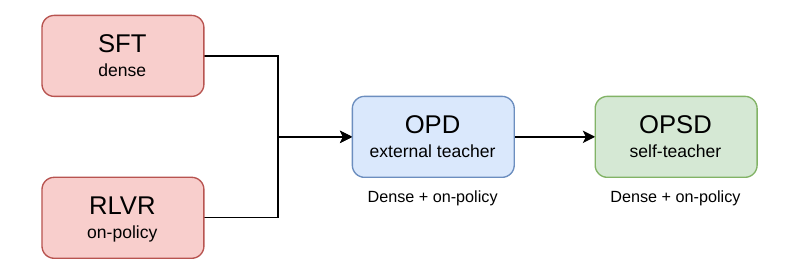}
    \caption{Genealogy of the training methods leading to OPSD.}
    \label{fig:carte}
\end{figure}

\bigskip
\noindent The three lineages converge into a single method.
\begin{itemize}
    \item From RL, OPSD keeps the \textit{on-policy} rollout, which avoids
    exposure bias;
    \item from distillation, it keeps the token-level \textit{density} of the
    signal, which addresses credit assignment;
    \item from privileged information, it keeps its independence from any
    external model (\textit{self-distillation}).
\end{itemize}
Figure~\ref{fig:carte} gives an overview of this lineage.

\bigskip

\medskip
\begin{axe}
\paragraph{\textcolor{accent}{Key takeaway --- Genealogy}}
Each method in the chain fixes the previous one's defect. SFT is dense but
off-policy. RL is on-policy but sparse. On-policy distillation is both, at the
price of a larger teacher. OPSD removes that price by replacing capability with
information: the teacher is the same model, better informed.
\end{axe}


\bigskip
\subsection{The OPSD Mechanism in Detail}
\label{sec:mecanisme}

\bigskip
\paragraph{Notation.} We use the following terms throughout:
\begin{itemize}
    \item $p_\theta$: a model (e.g. Qwen3 1.7B) with weights $\theta$, where:
    \begin{itemize}
        \item $p_S$: the student model;
        \item $p_T$: the teacher model.
    \end{itemize}
    \item $(x, y^\star)$: a pair drawn from the training set, where:
    \begin{itemize}
        \item $x$ is the prompt, i.e. the problem given to the model;
        \item $y^\star$ is the reference solution to problem $x$.
    \end{itemize}
    \item $V$: the vocabulary of model $p_\theta$, that is, the set of tokens $v$ it knows (for Qwen3, $\mathrm{card}(V) \approx 150{,}000$\footnote{The softmax runs over the model's logit dimension, $151{,}936$ for Qwen3; the tokenizer itself defines $151{,}669$ entries, the remainder being unused padding. We round to $150{,}000$ throughout for readability.}).
    \item $\hat y = (\hat y_1,\dots,\hat y_{\lvert\hat y\rvert})$: the rollout
    produced by the student. The index $n$ denotes a position within this rollout, to be distinguished from the index $v$, which denotes an item of the vocabulary $V$.
\end{itemize}

\begin{figure}[htbp]
    \centering
    \includegraphics[width=1\textwidth]{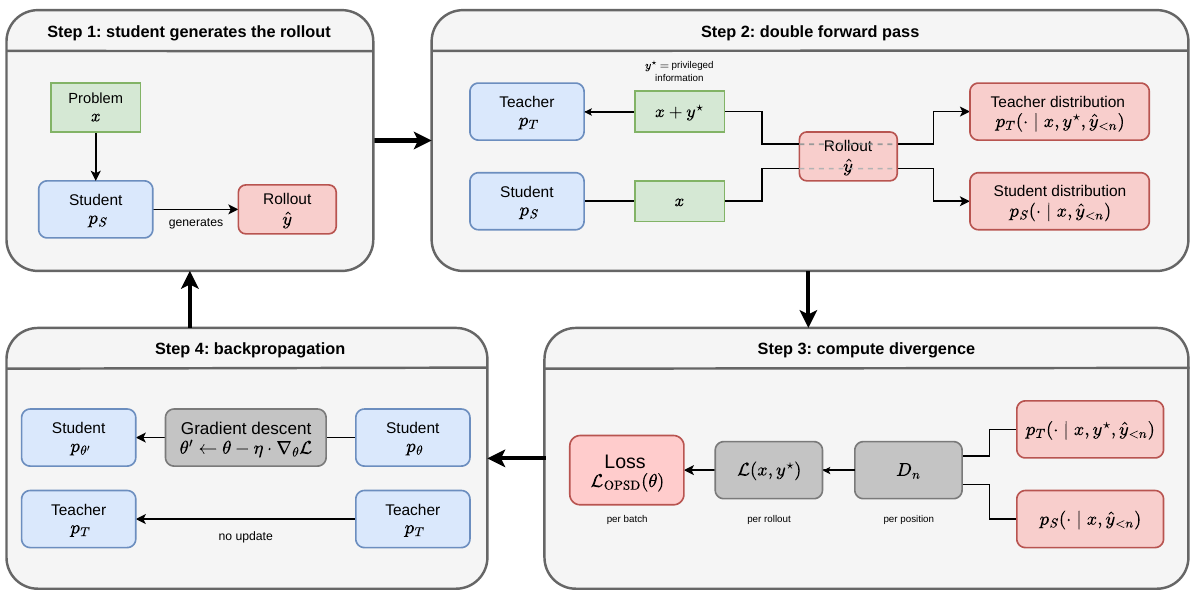}
    \caption{Overview of On-Policy Self-Distillation (OPSD).}
    \label{fig:schema-OPSD}
\end{figure}

\paragraph{Step 1: generating a rollout.} The prompt $x$ is passed to the student, which samples its answer autoregressively, token by token:
\[\hat{y} = (\hat{y}_1, \hat{y}_2, \dots, \hat{y}_{\lvert\hat{y}\rvert}) \sim
p_S(\cdot\mid x)\]
For each token $\hat{y}_n$, the procedure is as follows:
\begin{enumerate}
    \item the student $p_S$ predicts a probability distribution over $V$: every token $v$ known to the model is assigned a probability, conditioned on the preceding token sequence $\hat y_{<n}$;
    \item the model samples a token $\hat{y}_n$ from that distribution;
    \item the new token is appended to the context ($\hat{y}_{<n}$ becomes $\hat{y}_{<n+1}$);
    \item the operation is repeated for $\hat{y}_{n+1}$.
\end{enumerate}
In the OPSD paper, rollout length is capped at $1{,}024$ tokens. We now have the pair $(x, \hat{y})$, where $x$ is the problem and $\hat{y}$ the answer produced by the model.

\paragraph{Step 2: scoring the rollout.} The teacher then scores that rollout. For each token $\hat{y}_n$, it predicts a probability distribution conditioned on the preceding sequence $\hat y_{<n}$. The teacher generates nothing; it only scores the student's rollout token by token.
\begin{enumerate}
    \item the teacher is given the prompt $(x + y^\star + \text{``instructions''})$, made up of the problem and its solution. For instance: ``What is $3 \times 4$?'' + ``$12$'' + ``Having read the solution, produce your own reasoning to answer the problem.'';
    \item for each position $n$ of $\hat{y}$, the teacher's next-token distribution is collected:
    \[\forall n=1, \dots,\lvert\hat{y}\rvert, \qquad
    p_T(\cdot\mid x, y^\star, \hat{y}_{<n})\]
    \item in parallel, the student's distribution is collected as well:
    \[\forall n=1, \dots,\lvert \hat{y}\rvert , \qquad p_S(\cdot\mid x, \hat{y}_{<n})\]
    \item for a rollout of at most $1{,}024$ tokens and a vocabulary of $150{,}000$ tokens, this yields two probability matrices of roughly $1{,}024 \times 150{,}000$.
\end{enumerate}
Both distributions share the same prefix $\hat{y}_{<n}$. The teacher differs only in that it additionally holds the privileged information $y^\star$. This is also where OPSD gains an advantage over GRPO, since these operations are parallelizable: the distributions at all $n$ positions can be computed simultaneously. Only Step 1 is sequential.

\paragraph{Step 3: computing the loss.} With both matrices in hand, we measure the gap between the teacher's and the student's predictions at each position.
\begin{enumerate}
    \item at each position $n$, we compute the divergence $D_n$ (\textit{forward KL}) between the two distributions:
    \begin{align*}
        D_n &= \KL\left(p_T(\cdot\mid x, y^\star, \hat{y}_{<n})
        \,\middle\|\, p_S(\cdot\mid x, \hat{y}_{<n})\right) \\
        &= \sum_{v \in V}p_T(v)\log \left( \dfrac{p_T(v)}{p_S(v)} \right) \\
        &= \sum_{v \in V} l_{n,v}
    \end{align*}
    The scalar $D_n$ measures how large the teacher--student gap is at that position, obtained by summing the contribution of each vocabulary item $v$. One subtlety: before summing, \textit{clipping} is applied, with each dimension-wise contribution $l_{n,v}$ capped at $\tau$ to bound the influence of any single vocabulary item:
    \[ D_n^{\mathrm{clip}} = \sum_{v\in V} \min(l_{n,v}, \tau)\]

    \item the $D_n$ are averaged over the whole rollout:
    \[ \mathcal{L}(x,y^\star)=\dfrac{1}{\lvert \hat{y}\rvert }
    \sum_{n=1}^{\lvert \hat{y}\rvert }D_n^{\mathrm{clip}}\]
    This scalar measures the teacher--student gap over a complete rollout.
    \item in practice, several pairs $(x,y^\star)$ are processed before the weights are updated. Averaging the $\mathcal{L}(x,y^\star)$ yields the final loss:
    \[\mathcal{L}_{\mathrm{OPSD}}(\theta)=\dfrac{1}{\lvert B\rvert }
    \sum_{(x,y^\star)\in B} \mathcal{L}(x, y^\star)\]
    where $B$ is the batch of rollouts.
\end{enumerate}
We are left with a single scalar: how far the student is from the teacher. This is the gap we minimize, and the one we monitor during training.

\bigskip
The founding paper adopts the forward KL rather than the reverse KL. \citet{zhao2026opsd} report that the forward KL ``consistently yields the strongest gains'', the informed teacher serving as a reference distribution to be covered. This choice contrasts with the reverse-KL tradition of generative distillation \citep{gu2024minillm, lu2025blog}. The space of possible divergences, forward, reverse, or a JSD-style interpolation, is one of the axes reopened by the recent work we examine below (\S\ref{sec:A1}).

\paragraph{Step 4: computing the gradient.} We now have a signal that gives, at each position $n$ and for each vocabulary item $v$, the gap between teacher and student. Reducing $\mathcal{L}$ requires knowing how much to move each weight, that is, the gradient $\nabla_\theta\mathcal{L}$. The computation proceeds in stages:
\begin{enumerate}
    \item loss $\to$ distribution of $S$: we differentiate $\mathcal{L}$ with respect to $p_S(v)$, which gives a vector of dimension $\lvert V\rvert$ whose components indicate the direction in which to move;
    \item distribution $p_S \to$ logits: the $p_S$ come from a softmax over logits $z_n \in \mathbb{R}^{\lvert V\rvert}$. Composing the derivative of the KL with that of the softmax gives:
    \[\dfrac{\partial \mathcal{L}}{\partial z_{n,v}} \propto p_S(v)-p_T(v)\]
    from which the direction follows:
    \begin{itemize}
        \item $p_S(v) < p_T(v)$: negative derivative, so the logit is increased;
        \item $p_S(v) > p_T(v)$: positive derivative, so the logit is decreased.
    \end{itemize}
    Each vocabulary item $v$ therefore receives an instruction: up or down, and by how much;
    \item logits $\to$ weights $\theta$: the logits are the network's output. Backpropagation works back through the layers via the chain rule to the contribution of each weight $\theta$, giving:
    \[\nabla_\theta \mathcal{L} = \left[\dfrac{\partial \mathcal{L}}
    {\partial \theta_1}, \dfrac{\partial \mathcal{L}}{\partial \theta_2},
    \dots\right]\]
\end{enumerate}

\paragraph{Step 5: updating the weights.} The optimizer takes a gradient-descent step:
\[ \theta \leftarrow \theta - \eta \cdot \nabla_\theta \mathcal{L}\] 
where $\eta$ is the learning rate. Each weight moves slightly in the direction that locally reduces $\mathcal{L}$, bringing $p_S$ closer to $p_T$.

\bigskip
\noindent In summary:
\begin{enumerate}
    \item take a new pair $(x, y^\star)$;
    \item the student generates a rollout: $\hat{y} \sim p_S(\cdot \mid x)$;
    \item the teacher scores the rollout token by token:
    $p_T(\cdot\mid x, y^\star, \hat y_{<n})$;
    \item the token-level divergence between the two distributions is computed under the forward KL;
    \item the divergences are aggregated into a single scalar $\mathcal{L}$, which measures the size of the teacher--student gap;
    \item backpropagation is performed on the student only. The teacher is fixed.
\end{enumerate}

\bigskip
A second framework of the same kind appeared at the same time.
\citet{hubotter2026sdpo} introduced \textbf{Self-Distillation Policy Optimization} (SDPO), which shares the founding intuition of OPSD: a self-teacher informed by privileged information provides a dense signal to the student, with neither an external teacher nor a reward model. SDPO changes the nature of that information. Where OPSD conditions its teacher on the reference solution $y^\star$, SDPO conditions it on \emph{textual feedback from the environment}: execution error messages, the output of a verifier, or the assessment of a judge. For a coding problem:
\begin{enumerate}
    \item the student samples a rollout $\hat{y}$ from the problem $x$;
    \item the code is executed, and the environment returns textual feedback $f$, for instance the trace of a runtime error;
    \item the rollout is re-scored under a self-teacher conditioned on this feedback, $p_T(\cdot \mid x, f, \hat{y}_{<n})$;
    \item the teacher's corrected next-token distribution is distilled into the student's policy.
\end{enumerate}
The method exploits the model's ability to identify its own mistakes in hindsight. Once the error is known, the teacher can correct the student's tokens so that they are avoided in later generations. One further difference concerns the teacher itself. OPSD keeps it frozen at the initial policy, whereas SDPO regularizes it for stability, either through an exponential moving average (EMA) of the student's weights or through interpolation with the initial teacher. We return to teacher dynamics in \S\ref{sec:A4}.

\bigskip
The two methods therefore belong to the same family: on-policy self-distillation guided by privileged information (PI). They differ, however.
\begin{enumerate}
    \item \textbf{The nature of the PI}: the reference solution $y^\star$ for OPSD, execution feedback $f$ for SDPO. SDPO is thus naturally suited to domains with a verifiable environment (code, tool use), whereas OPSD presupposes a dataset of annotated solutions.

    \item \textbf{Teacher stabilization}: OPSD keeps its teacher frozen at the initial policy, while SDPO lets it evolve under regularization.

    \item \textbf{Scope}: SDPO can also be applied at test time to a single hard question, by iteratively distilling feedback into the policy, a regime OPSD does not explore.
\end{enumerate}

\medskip
\begin{axe}
\paragraph{\textcolor{accent}{Key takeaway --- OPSD mechanism}} OPSD is a loop: the student generates, the teacher (the same model $+\,y^\star$) scores each token under the forward KL, and backpropagation is applied to the student only. The teacher stays fixed. SDPO is the same machinery, but the privileged information is execution feedback.
\end{axe}


\bigskip
\subsection{Strengths, Weaknesses and Open Problems}
\label{sec:tensions}

\bigskip
In its founding paper, OPSD delivers on part of its promise. Density, however,
raises a problem: it increases the risk of collapse.

\paragraph{What OPSD brings.} The figures below come from the founding paper
\citep{zhao2026opsd}. They should be read in view of \S\ref{sec:eval}, a
reminder of how fragile these benchmarks are.

\begin{itemize}
    \item \textbf{Efficiency.} Where GRPO samples 8 rollouts of up to 16k
    tokens per problem, OPSD makes do with a single generation capped at
    $1{,}024$ tokens. At comparable performance on mathematical reasoning, it
    consumes far fewer generated tokens per problem. The gain does not
    translate into a compute gain, however. An OPSD optimization step requires
    two forward passes and one backward pass, against a single backward pass
    for GRPO. At equalized budget, one OPSD step costs roughly twice a GRPO
    step ($20.6$\,s against $11.2$\,s on Qwen3-8B, 8$\times$H100)
    \citep{li2026localizing}. The advantage is faster convergence in number of
    steps, not a lower unit cost. A run on Qwen3-1.7B completes in about
    fifteen minutes on 4~H100
    GPUs.\footnote{\href{https://github.com/siyan-zhao/OPSD}{github.com/siyan-zhao/OPSD}}

    \item \textbf{Performance.} Despite this reduced budget, OPSD matches or
    exceeds GRPO on mathematical reasoning, and outperforms off-policy
    distillation. The choice of the \textit{forward KL} is decisive here: the
    authors report a rise from $36.7$ to $43.9$ on AIME25 by
    step~50.\footnote{Best reported scores (Table~2 of the paper) for Qwen3-1.7B with OPSD: $57.2\%$ on AIME24, $43.9\%$ on AIME25, $29.2\%$ on HMMT25. These are best-over-checkpoints figures; on AIME25 the end-of-training value is $41.1\%$. Per \S\ref{sec:eval}, checkpoint selection inflates such numbers.}

    \item \textbf{Autonomy.} The teacher is the model itself, so no larger
    model is required.
\end{itemize}

\noindent SDPO (\S\ref{sec:mecanisme}) confirms that the family transfers
beyond mathematics, to code and agentic tasks, with efficiency gains of the
same order.

\bigskip
Density accelerates learning. It also accelerates the model's drift toward its
own biases. Work extending OPSD measures degradations of up to $-17\%$ (avg@16)
on \textit{thinking} models \citep{kaur2026rethinking}, with comparable effects
out of domain \citep{kim2026why}. Dense supervision can narrow the diversity of
reasoning, or the entropy of the policy, down to the fixed point at which
teacher and student coincide. This is \textbf{collapse}, the symptom the rest of
the paper seeks to control.

\bigskip
\noindent Three \textbf{levers} act on it:
\begin{itemize}
  \item \textbf{Lever A. Signal geometry.} Which divergence, and how dense? The
  choice determines whether the student covers the teacher's behaviours or
  locks onto one of them. A denser signal is not always preferable.
  \item \textbf{Lever B. Privileged information.} Which information should the
  teacher be given? Too informative, and it biases the student, which then
  memorizes shortcuts unavailable at test time.
  \item \textbf{Lever C. Loop stability.} The teacher is the model itself, so
  the loop can drift. Its update rule, the forgetting of earlier capabilities,
  and the scheduling of guidance all bear on stability.
\end{itemize}
The levers are not independent. The choice of privileged information bears on
all three, which is why it occupies a central place here. Part 2 takes up the
symptom and each lever in turn.

\medskip
\begin{axe}
\paragraph{\textcolor{accent}{Key takeaway --- Strengths \& weaknesses}}
OPSD matches GRPO on mathematical reasoning while generating far fewer tokens,
and it needs no larger model. Its strength is density, and density is also its
main danger: it heightens the risk of collapse. Three levers can offset that
risk: the geometry of the signal, the choice of privileged information, and the
stability of the loop.
\end{axe}

\bigskip
\subsection{Evaluating These Models}
\label{sec:eval}

On small models and reasoning benchmarks, performance measurements are fragile in ways that are now well documented.

\paragraph{Three sources of illusion.}
\begin{itemize}
    \item \textbf{Variance.} A competition benchmark such as AIME comprises only thirty questions. A single question flipping shifts the score by more than three points, and the spread between two decoding seeds can reach fifteen
    \citep{hochlehnert2025sober}. A ``$+3$ points'' from a single decode is usually noise.

    \item \textbf{Contamination.} AIME~2024 problem statements are partly present in pre-training data, to the point that some models complete half
    of them from memory while failing on benchmarks released after their training cutoff \citep{wu2026reasoning, matharena2025}.

    \item \textbf{Model-family specificity.} On Qwen models, even a random training signal can raise the score. The effect is absent on Llama and OLMo, and comes from pre-training rather than from the method under
    evaluation \citep{shao2025spurious, octothinker2025}.
\end{itemize}

\paragraph{What a rigorous reading requires.} These pitfalls yield the grid we apply throughout Part 2.
\begin{itemize}
    \item On the measurement side, a single score is not enough. We look for an average over several samples and several seeds (avg@$k$), with a confidence interval.

    \item We also track pass@$k$, which exposes a loss of diversity that a mean score conceals \citep{yue2025rl}, and G-Pass@$k$, which measures the stability of reasoning beyond its one-off success \citep{liu2025gpassk}.

    \item On the protocol side, four controls separate signal from artefact: a comparison against null or random privileged information; a comparison at equalized compute budget; a contamination test contrasting older and more recent benchmarks; and a replication outside the Qwen family.
\end{itemize}

\medskip
\begin{axe}
\paragraph{\textcolor{accent}{Key takeaway --- Evaluating models}} On small models, reasoning scores are fragile: benchmark variance, contamination, and effects specific to the Qwen family. No figure in Part 2 should be read without checking how it was obtained, namely how many seeds, whether pass@$k$ is reported, and which model family was used.
\end{axe}

\bigskip
\section{Developments Since the Founding Paper}
\label{sec:partie2}

Each subsection below follows the same pattern: where the field stood at the founding paper, what it has produced since, and what remains open.


\subsection{Lever A --- Signal Geometry: Which Divergence, Which Density?}
\label{sec:A1}

The first tension concerns the shape of the distillation signal. It covers two
coupled choices: the \textbf{direction of the divergence} that brings student
and teacher together, and the \textbf{density} of that signal, meaning the
number and relative weight of the supervised tokens. Both involve the same
trade-off: gaining performance without collapsing diversity.

\paragraph{The direction of the divergence.} The space of possible divergences
was already framed by GKD \citep{agarwal2024gkd} in 2023, which allows the
forward KL, the reverse KL, or their interpolation (JSD)
interchangeably.\footnote{\href{https://hiroakih.me/kl-divergence.html}{hiroakih.me/kl-divergence.html}:
an interactive page that helps build intuition for the behaviour of the
different divergences.} The same work compares them on translation,
summarization and arithmetic tasks. The \textit{reverse KL} achieves the best
performance and the lowest diversity, the \textit{forward KL} the reverse, and
JSD sits between the two. MiniLLM \citep{gu2024minillm} popularizes the reverse
KL for LLMs at the same time. By pushing the student onto the teacher's
dominant modes, it prevents the student from overestimating low-probability
regions and improves calibration.

The reverse KL therefore looks like the most attractive option for LLM
post-training. Its drawbacks appear once it is observed over several attempts.
DPH-RL \citep{li2025dphrl} shows that it \textbf{accelerates diversity
collapse}: pass@1 rises while pass@$k$ falls, with no safeguard against the
model drifting away from its knowledge base. The phenomenon worsens on
out-of-domain tasks. The direction of the divergence is therefore a first-order
lever. It justifies OPSD's choice of the forward KL, and the existence of
stabilized variants such as the skew KL of DistiLLM \citep{ko2024distillm}.

\paragraph{Density and its price.} The second choice concerns the quantity of
signal. The intuition that ``denser is better'' is directly contradicted by the
most recent work. Denser $\neq$ Better \citep{wang2026denser} establishes that
\textbf{density is a powerful but fragile signal}. Distilling the full chain of
thought helps on tasks with short traces, such as tool use, but degrades
mathematics and science, whose long traces tend to surface artefacts. In
continual learning, SDPO specializes quickly and then collapses, whereas
sparse-reward RL of the GRPO kind retains more.

Unmasking OPD \citep{armandpour2026unmasking} explains that fragility. By
comparing the distillation gradient to an ideal per-token gradient, the authors
measure an \textit{alignment score}: positive when the teacher pushes towards
success, null when the signal is spent on style, negative when it pushes
towards failure. Their findings:
\begin{itemize}
    \item distillation helps \textbf{mainly on erroneous trajectories}; when
    the student is already on the right track, the teacher is little more than
    a noisy signal;
    \item the best teacher depends on the \textbf{student's capacity}: on a
    0.6B model, self-distillation is two to three times better than an external
    teacher, an advantage that does not carry over to a 1.7B model;
    \item distilling only on positively aligned tokens, roughly half the total,
    would improve the signal by a factor of ten to fifteen.
\end{itemize}
Uniform density therefore lets stylistic tokens and artefacts dilute, and even
corrupt, the useful signal.

\paragraph{Towards selective density: token weighting.} This diagnosis points
to the most promising direction in the section: make density
\textbf{selective}, by weighting tokens according to their importance.

Entropy-Aware OPD \citep{entropyaware2026} gives a concrete example. Standard
on-policy distillation relies on the reverse KL, which is \emph{mode-seeking}:
it pushes the student to imitate the teacher's most confident predictions. This
works well when the teacher is sure of itself, and becomes unstable when it is
not, that is, on \textbf{high-entropy} tokens. Yet these are the decisive
tokens: those at which reasoning branches and several continuations remain
plausible. Forcing the student onto a single choice there crushes its
diversity. The authors measure the effect. On high-entropy tokens, the student's most
probable token changes $84$ times over the course of training, against only $7$
times in the low-entropy regime. The student maintains only $6.8\,\%$
high-entropy tokens where the teacher retains $18.5\,\%$. The student becomes
impoverished exactly where it ought to explore.

Their remedy is simple: \textbf{keep the reverse KL everywhere, but add a
forward KL on the teacher's high-entropy tokens only}. Where the teacher is
confident (low entropy), the reverse KL suffices and the student imitates the
right token. Where the teacher is uncertain (high entropy), the forward KL,
being \emph{mode-covering}, forces the student to cover the full range of
continuations the teacher deems plausible, instead of collapsing onto a single
one. The result combines the precision of imitation where it is reliable with
the robustness of coverage where the signal is ambiguous. The compute overhead
is about $4.5\,\%$ per step. The clipping used in OPSD (\S\ref{sec:mecanisme})
is a crude precursor: it bounds the contribution of each vocabulary item, which
mechanically attenuates formatting positions. It never distinguishes positions
according to the teacher's uncertainty.

DPH-RL \citep{li2025dphrl} applies a related logic at a different level. It is
not a distillation method: it is an RL method of the GRPO kind that rethinks
the role of the divergence term. Selectivity therefore operates problem by
problem rather than token by token. Before training, the dataset is partitioned
in two: the problems the base model can already solve, and the rest. The
divergence then varies with the nature of the problem:
\begin{itemize}
    \item \textbf{mastered problems}: a \emph{mass-covering} divergence
    (forward KL or JS) is added, anchoring the model to its initial policy.
    This amounts to having it revise what it already knows so it does not
    forget, a \emph{rehearsal} mechanism against catastrophic forgetting;
    \item \textbf{unmastered problems}: the divergence is removed entirely and
    the model explores freely, guided by the reward alone. A skill that has not
    yet been acquired cannot be revised.
\end{itemize}
Anchoring is done towards the frozen initial policy, which makes the method
efficient: no reference model needs to run online during training. The authors
propose two variants, according to the divergence used on mastered problems:
DPH-F (forward KL) and DPH-JS (Jensen--Shannon). They recommend the latter. The
JS divergence provides a more flexible anchor, symmetric and more stable, which
preserves diversity without imposing the rigid memorization that the forward KL
would entail.

\paragraph{What remains open.}
\begin{itemize}
    \item \textbf{A weighting criterion that is both justified and computable.}
    The two available criteria have symmetric defects. Alignment with the ideal
    gradient \citep{armandpour2026unmasking} is the better founded, but it is
    measured after the fact, since the outcome of the trajectory must be known
    before a token can be said to have pushed towards success. Teacher entropy
    \citep{entropyaware2026} is available online, at every step, but nothing
    guarantees that it correctly approximates alignment. A criterion that is
    both available during training and correlated with a token's actual
    usefulness remains to be built.

    \item \textbf{The granularity of the weighting.} Selectivity is applied
    today either token by token or problem by problem \citep{li2025dphrl}.
    Nothing indicates that these are the optimal scales, nor that the same
    granularity suits short traces and long reasoning chains.

    \item \textbf{An announced gain that has yet to be demonstrated.}
    \citet{armandpour2026unmasking} estimate that distilling only on positively
    aligned tokens would improve the signal by a factor of ten to fifteen. This
    is an oracle measurement, obtained outside training, that no method has yet
    converted into an effective gain.
\end{itemize}

\medskip
\begin{axe}
\paragraph{\textcolor{accent}{Key takeaway --- Signal geometry}}
Two coupled choices shape the signal, and neither is a secondary setting. The
direction of the divergence decides whether the student covers the full range
of the teacher's behaviours or settles on one of them, which makes it the
control on the performance--diversity trade-off. The distribution of density is
more counter-intuitive: a rollout contains only a handful of decisions that
genuinely commit the reasoning, so uniform weighting lets formatting occupy
most of the gradient. The lever is thus not the quantity of signal, but its
selectivity.
\end{axe}

\bigskip
\subsection{The Symptom --- Collapse: Two Families of Causes}
\label{sec:A2}

\bigskip
The downside of density is the pathology most feared in OPSD:
\textbf{collapse}. The term denotes the progressive narrowing of the set of reasoning paths the model is able to produce. It shows at three levels.
\begin{itemize}
    \item In \textbf{behaviour}, diversity collapses: pass@1 rises, but pass@$k$ flattens or even declines. The model succeeds more often, yet loses the ability to explore rare but correct solutions. It performs better on problems of the kind seen in training, and worse out of domain.
    
    \item In the \textbf{token distribution}, entropy tends towards zero: the model concentrates its probability mass on a decreasing number of continuations.
    
    \item In \textbf{geometry}, the target becomes unimodal: the model closes in on a single mode. For a given problem, it learns one correct solution and uses only that one, without exploring alternatives.
\end{itemize}
These three levels are often presented as equivalent. They are not. \citet{nicolicioiu2026} measure, on Qwen3-8B, that self-distillation raises pass@1 from $71.9$ to $73.4$ while pass@16 falls from $83.6$ to $78.5$: mean success improves, functional diversity recedes. They further observe that this same model displays a token entropy \emph{higher} than that of the GRPO-trained model, even though its functional diversity is \emph{lower}. Entropy is therefore not a valid proxy for diversity, which is why the grid of \S\ref{sec:eval} requires pass@$k$ rather than the mean score alone.

\bigskip
Two families of causes coexist. The first predates OPSD and is found in any RL method, as in any self-training loop. The second is specific to OPSD and stems from conditioning the teacher on privileged information. The distinction matters: only the second depends on what the teacher is given, and only the second offers OPSD a lever.

\paragraph{The collapse that predates OPSD.} Neither mechanism here involves a teacher or privileged information.

The first lies in the RL gradient itself. \citet{cui2025entropy} give its law, $R = -a\,e^{H} + b$: performance is bounded by an exhausted entropy budget, and that budget declines monotonically over training. Each update that favours one correct answer raises its probability and lowers that of other answers, equally correct but slightly less likely. The mechanism operates within a single training run, and it follows from the objective being optimized, which rewards success without ever valuing exploration.

The second is \emph{model collapse}, which \citet{shumailov2024nature} describe for any loop in which a model is retrained on its own generations. It unfolds in two stages. The tails of the distribution disappear first, meaning the rare but valid solutions. Convergence towards a single mode of near-zero variance follows. Unlike the previous mechanism, it takes place from one generation to the next, and its cause is finite sampling rather than the optimization objective. The signature it describes, tails first and mode second, matches what is observed in OPSD, but its mechanism transfers imperfectly.
\citet{gerstgrasser2024accumulate} show that collapse presupposes that synthetic data \emph{replace} real data, and that merely accumulating the two bounds the error. OPSD does re-inject real data at every step: the reference solution $y^\star$. It enters through the teacher's conditioning, however, rather than through the student's training distribution. Model collapse therefore describes the shape of the phenomenon, but not its cause.

\paragraph{The mechanism specific to OPSD: PMI.} The narrowing might be attributed to the direction of the divergence, but OPSD adopts the forward KL, which is mass-covering and preserves coverage. The cause therefore appears to be finer, and to operate at the level of individual tokens.
\citet{antisd2026pmi} show that the signal transmitted token by token from teacher to student is a \textit{pointwise mutual information} (PMI) between the token produced and the privileged context. Conditioning the teacher on the solution turns it into an oracle: it strongly rewards the tokens that the solution already entails, such as connectives and verifiable content, and penalizes deliberation tokens (``wait'', ``let'', ``maybe''), which an oracle no longer needs since it knows the answer. Yet this deliberation phase is what enables the student to conduct multi-step search at inference time. Three works confirm the mechanism from different angles.
\begin{itemize}
    \item \citet{kim2026why} call it the \emph{suppression of epistemic verbalization}: an over-informed teacher expresses less uncertainty, the student loses it, and out-of-domain performance collapses.
    
    \item \citet{nicolicioiu2026} give its dynamics, \emph{rich-get-richer}: the demonstration sampled and given to the teacher as privileged information is most often the dominant mode, so that rare but correct strategies receive a weak signal and die out.
    
    \item \citet{kaur2026rethinking} localize it: privileged context lowers the \emph{fork rate}, that is, the proportion of decision points at which reasoning can change direction.
\end{itemize}
PMI may therefore be more than an explanation of the phenomenon. It suggests a predictive grid: the more directly the privileged information entails the tokens of the solution, the more the signal should inflate shortcuts and crush deliberation. On this reading, the severity of collapse depends on what the teacher is given, which makes the nature of the privileged information the central variable to control (\S\ref{sec:A3}).

\paragraph{Remedies.} The remedies proposed to date intervene neither at the same point nor on the same family of causes.
\begin{itemize}
    \item \textbf{On the RL gradient.} Rare tokens are protected by raising the clipping bound (Clip-Higher, in DAPO \citep{dapo2025}), or by targeting tokens with a high covariance between probability and logit update \citep{cui2025entropy}. These remedies address entropy collapse, hence the general cause, and apply to OPSD as to any RL method.

    \item \textbf{On the divergence.} A \emph{mode-seeking} objective is replaced by a \emph{mass-covering} divergence that preserves coverage. DPH-RL \citep{li2025dphrl} does so problem by problem rather than token by token, anchoring the model to its initial policy on the problems it already masters (\S\ref{sec:A1}). This remedy too targets the general cause: the loss of coverage.

    \item \textbf{On the sign of the signal.} The update is reversed where it does harm. Anti-SD \citep{antisd2026pmi} replaces gradient descent towards the teacher with a divergence \emph{ascent}, to encourage the model to explore rather than concentrate its probability mass. Of the three, it is the only one that targets the PMI mechanism directly.
\end{itemize}
Two of these three families address the general cause, only one the mechanism specific to OPSD. All of them intervene downstream, once the teacher has already been conditioned. None touches the variable that sits upstream: the information given to the teacher.

\medskip
\begin{axe}
\paragraph{\textcolor{accent}{Key takeaway --- Collapse}}
Collapse covers two families of causes. The first predates OPSD: the RL gradient erodes entropy, and any self-training loop impoverishes the tails of the distribution. The second is specific to OPSD: conditioning the teacher on the solution turns it into an oracle, which inflates the tokens the solution already entails and penalizes those of deliberation. PMI may be more than a post-hoc explanation. It suggests a grid for predicting which privileged information will collapse the student, though no study has yet tested it as a predictor.
\\[2pt]
Several remedies exist, none has reached consensus, and all of them act downstream of the teacher. The question they leave open is the one the next section takes up: which privileged information should the teacher be given, so as to guide it without crushing deliberation?
\end{axe}

\bigskip
\subsection{Lever B --- The Nature of the Privileged Information}
\label{sec:A3}

\bigskip
Collapse depends on \emph{what} the teacher is given. Which privileged
information should be chosen?

\paragraph{An old question.} Having a student learn with the help of
information only the teacher holds was theorized as early as 2009.
\citet{vapnik2009lupi} draw from it a rule that still holds: this help serves
to \emph{learn} better, not to be \emph{copied}. In their model, the privileged
information never enters the final decision; it serves only to identify which
examples are difficult. \citet{lopezpaz2016unifying} then show that
distillation is a special case of this framework: a teacher that distils in
effect transmits privileged information to its student. OPSD is its direct
descendant.

\paragraph{The lesson from robotics.} Robotics has already answered a closely
related question: when does privileged information help without doing harm?
Privileged information is safe if the student can reconstruct it on its own
from what it perceives at test time, and toxic if it must presuppose or
memorize it. Three results ground this criterion.
\begin{enumerate}
    \item The case that works. In \emph{Learning by Cheating}
    \citep{chen2019cheating}, an autonomous car is trained in two stages. A
    first agent ``cheats'': it sees the exact layout of the scene and learns to
    drive. A second agent, equipped only with a camera, imitates it. This works
    because the student can recover from the image what the teacher knew.

    \item The case that fails. \citet{weihs2021advisor} show that when the
    teacher acts on information unavailable to the student, that information is
    marginalized during imitation, producing an \emph{imitation gap} and
    provably poor policies. This is the ``presupposed'' or ``memorized'' case:
    the student is asked to reproduce a behaviour it cannot justify from its
    own observations.

    \item The right design. In RMA \citep{kumar2021rma}, the privileged
    information is never copied. The student learns to \emph{regenerate} it
    itself, from its own history. The target therefore remains reachable.
\end{enumerate}
Privileged information can help the student, and can equally harm it by
introducing data the student cannot access and that disrupt it at test time.
The choice therefore matters for how the student performs under deployment
conditions. This lineage is largely absent from the OPSD literature, which has
rediscovered its vocabulary without inheriting its results.

\paragraph{The same lesson, on the LLM side.} Work from 2026 recovers this gap
on LLMs. The central result is that of \citet{kaur2026rethinking}. They give
privileged information to the teacher and measure the effect on
\emph{thinking} models. Rather than helping, the privileged information
\textbf{degrades} these models, by up to $-17\,\%$ in relative terms (avg@16).
The explanation aligns with the PMI mechanism of \S\ref{sec:A2}. A teacher that
already knows the answer stops hesitating. It produces fewer deliberation
tokens (``wait'', ``maybe'', backtracking), and pushes the student to abandon
them. Yet these are the tokens that serve to explore several paths at inference
time. The student thus learns to skip a reasoning stage that is crucial at test
time. Two of their conclusions are decisive.
\begin{itemize}
    \item The effect depends on the model: the same information harms thinking
    models but helps instruction-tuned ones.
    \item The effect depends on the \emph{quantity} of information. A full
    demonstration (reasoning plus answer) yields the best gains when the
    generation budget is short. As the budget lengthens, the effect reverses,
    whereas the final answer alone keeps the model close to its base. Giving
    the teacher more is therefore not uniformly worse, only less stable.
\end{itemize}
The harm comes from the \emph{nature} of the privileged information, not from
self-distillation itself. A final result confirms this on an apparently
innocuous choice. \citet{nicolicioiu2026} give the teacher a correct
demonstration, sampled at random from among the student's successes. That
choice carries a hidden cost: already frequent solutions become even more
probable, and rare but correct ones disappear. This is the
\emph{rich-get-richer} effect of \S\ref{sec:A2}. In practice, the
demonstrations must be diverse as well as correct.

\paragraph{A taxonomy ordered by risk.} The safety criterion allows privileged
information to be ranked: the more an item of information \emph{presupposes}
the solution, the riskier it is, since the student may memorize it rather than
learn from it. The ranking below is only partially supported by the literature
and may therefore contain errors.
\begin{itemize}
    \item The \textbf{final answer} (oracle) is the riskiest. The student
    cannot reconstruct it, and the teacher has nothing left to deliberate
    about.

    \item The \textbf{worked solution} is the reference derivation together with its final answer, and it serves as the privileged information in standard OPSD. It presupposes the answer just as fully as the oracle does, but it also supplies the reasoning that leads there. It is therefore not simply ``more CoT’’.

    \item The \textbf{full chain of thought} (CoT) is presupposing in a weaker
    sense: it exhibits the reasoning but need not state the final answer. Some
    work proposes showing the teacher only the first half of the CoT, and finds
    this preferable to the whole.

    \item The \textbf{plan}, or skeleton of steps, gives the structure without
    the values. The target is more reachable.

    \item The \textbf{rubric} lists the criteria of a good answer without
    imposing a specific path. The teacher can then cover several lines of
    reasoning, so diversity is preserved.

    \item \textbf{Error feedback} and \textbf{action-only} information (the
    actions of a strong model, without its reasoning) are partial and
    conditional. These are the least presupposing.
\end{itemize}

\paragraph{The first controlled comparisons.} This ordering is no longer merely
hypothetical. Two works compare several kinds of privileged information at
fixed model and data.

\citet{kara2026alignment} contrast three self-teacher contexts under strict
self-distillation: a binary reward (GRPO), the reference solution (standard OPSD), and a step-aligned critique. The last is feedback
generated by a critic model that copies the correct steps of the student's
reasoning verbatim and rewrites only the incorrect ones, in the student's own
style, concentrating the learning signal on the tokens where the reasoning
fails. The aligned critique wins by $+5.27$ over OPSD conditioned on the
reference solution and by $+16.11$ over GRPO (avg@12). Their per-token
advantage analysis uncovers a second mechanism, distinct from PMI. When the
model sees the reference solution, it modifies its behaviour at every token,
including those already correct. The aligned critique modifies only the
incorrect ones, which makes the learning signal far more targeted.

\citet{dopd2026} broaden the comparison to five forms: the final answer,
step-wise hints with execution, step-wise hints without execution, summarized
hints, and no privileged information at all. The final answer falls below the
no-privilege baseline ($59.5$ against $63.0$ on C-Eval), whereas step-wise
hints without execution clearly dominate ($71.3$). Their conclusion meets the
safety criterion: what makes privileged information effective is not the
correctness of the answer it contains, but its capacity to transmit a skill.

These two results confirm both ends of the hierarchy above. The oracle is the
least transferable choice, harmful outside self-distillation and inert within
it, while intermediate abstraction appears to be the best option. They do not
corroborate it perfectly, however: the first compares only three forms, and the
second compares five but in a strong-to-weak distillation regime.
Table~\ref{tab:pi} summarizes the kinds of privileged information tested to
date.

\begin{table}[tbp]
\centering
\footnotesize
\setlength{\tabcolsep}{7pt}
\renewcommand{\arraystretch}{1.3}

\caption{\textbf{Not all privileged information is equal, and the kind easiest
to obtain is the least useful.} Entries are ordered by broadly decreasing presupposition of the
solution; the first two presuppose the answer equally and differ in whether the derivation is supplied. The \emph{Regime} column distinguishes strict
self-distillation (the OPSD framework) from distillation of a strong model into
a weak one (the OPD framework). The \emph{Requires} column indicates the
resources needed to construct the privileged information.}
\label{tab:pi}
\medskip

\begin{tabularx}{\textwidth}{@{}
  X
  >{\centering\arraybackslash}p{2.5cm}
  >{\raggedright\arraybackslash}p{2.5cm}
  >{\raggedright\arraybackslash}p{2.5cm}
  >{\raggedright\arraybackslash}p{2.5cm}
  @{}}
\toprule
\textbf{Nature of the PI} &
\textbf{Reconstructible}\newline\textbf{at test time} &
\textbf{Measured effect} &
\textbf{Regime tested} &
\textbf{Requires} \\
\midrule

Final answer only\newline\citep{dopd2026, kaur2026rethinking} &
no &
harmful or inert\,\textcolor{blue}{$\mathbf{^{(a)}}$} &
strong $\to$ weak; self-dist. &
annotated solutions \\
\addlinespace[0.45em]

Solution: reasoning + answer (OPSD) \citep{zhao2026opsd} &
no &
harmful on \emph{thinking}\,\textcolor{blue}{$\mathbf{^{(b)}}$} &
self-dist. &
annotated solution \\
\addlinespace[0.45em]

Partial trace, anchor\newline(AR-OPD) \citep{aropd2026} &
partial &
positive\,\textcolor{blue}{$\mathbf{^{(c)}}$} &
self-dist. &
annotated traces \\
\addlinespace[0.45em]

Plan, hints\newline(DOPD) \citep{dopd2026} &
yes &
positive\,\textcolor{blue}{$\mathbf{^{(d)}}$} &
strong $\to$ weak &
annotated solutions + generation \\
\addlinespace[0.45em]

Error-aligned\newline critique \citep{kara2026alignment} &
yes &
positive\,\textcolor{blue}{$\mathbf{^{(e)}}$} &
self-dist. &
critic model \\
\addlinespace[0.45em]

Rubric\newline\citep{rubric2026} &
yes &
diversity yes, accuracy little\,\textcolor{blue}{$\mathbf{^{(f)}}$} &
self-dist. &
manual authoring \\
\addlinespace[0.45em]

Execution feedback\newline(SDPO) \citep{hubotter2026sdpo} &
yes &
positive\,\textcolor{blue}{$\mathbf{^{(g)}}$} &
self-dist. (code) &
verifiable environment \\

\bottomrule
\end{tabularx}

\vspace{0.7em}
\begin{minipage}{\textwidth}
\scriptsize
\raggedright
\textbf{Notes.}
\textcolor{blue}{$\mathbf{^{(a)}}$}~In the strong~$\to$~weak regime, the final answer scores $59.5$ on C-Eval, \emph{below} the $63.0$ obtained with no privileged information at all \citep{dopd2026}. Under strict
self-distillation, it instead keeps the student close to the base model, neither degrading nor improving it \citep{kaur2026rethinking}.
\textcolor{blue}{$\mathbf{^{(b)}}$}~Up to $-17\,\%$ in relative terms (avg@16)
across five \emph{thinking} models \citep{kaur2026rethinking}.
\textcolor{blue}{$\mathbf{^{(c)}}$}~An anchor built on the first half of the
trace reduces \emph{shortcut events} by more than $20\,\%$ \citep{aropd2026}.
\textcolor{blue}{$\mathbf{^{(d)}}$}~Step-wise hints without execution reach
$71.3$, against $63.0$ with no privileged information, across five forms compared at fixed model and data \citep{dopd2026}.
\textcolor{blue}{$\mathbf{^{(e)}}$}~$+5.27$ over the teacher conditioned on the reference solution and $+16.11$ over GRPO (avg@12) \citep{kara2026alignment}.
\textcolor{blue}{$\mathbf{^{(f)}}$}~Diversity is preserved and entropy rises
even under the forward KL, but accuracy gains remain modest, and a hand-written rubric clearly outperforms a model-generated one \citep{rubric2026}.
\textcolor{blue}{$\mathbf{^{(g)}}$}~$48.8\,\%$ against $41.2\,\%$ for GRPO on LiveCodeBench v6; at test time, the same discovery probability as best-of-$k$
sampling is reached with three times fewer attempts
\citep{hubotter2026sdpo}.
\end{minipage}
\end{table}

\paragraph{From diagnosis to remedies.} Recent work no longer merely observes the problem. It names, measures and corrects the \emph{leakage} of privileged information: the shortcuts the student learns during training but will not find
at test time. Four strategies stand out.
\begin{itemize}
    \item \textbf{Decomposing the target.} AR-OPD \citep{aropd2026} splits the
    teacher's signal in two. The \emph{anchor} comes from showing the teacher
    only the first half of the trace, without the answer, and gives a target
    the student can reach. The \emph{residual} is the difference between the
    full oracle target and that anchor, and it carries the leakage. Only a
    fraction $\lambda$ of the residual is retained:
    \[
        q_\lambda = p_T^{\text{anchor}} + \lambda \,\big(p_T^{\text{oracle}} -
        p_T^{\text{anchor}}\big), \qquad \lambda = 0.6 .
    \]
    \emph{Shortcut events} then fall by more than $20\,\%$. The question
    ``which privileged information?'' becomes ``how much of it remains
    learnable at the token level?''.

    \item \textbf{Purifying the signal.} Any privileged information mixes two
    things: a \emph{transferable} signal (understanding the problem) and a
    \emph{non-transferable} one (a shortcut specific to the reference answer).
    To separate them, Purified OPSD \citep{purifiedopsd2026} conditions a
    teacher on the answer \emph{without} the problem. That teacher reveals what
    comes from the shortcut rather than from the reasoning, and that share is
    then removed. The student's uncertainty markers remain stable as a result.

    \item \textbf{Changing the form of the information.} Rather than the exact
    solution, rubric-based distillation \citep{rubric2026} gives the teacher
    the \emph{criteria} of a good answer. The teacher no longer designates a
    single path; it validates several lines of reasoning. Diversity is
    preserved, and entropy rises even under the forward KL. Accuracy gains
    remain very modest, however, and a hand-written rubric performs far better
    than a model-generated one, which raises the question of how this method
    would scale.

    \item \textbf{Modulating confidence token by token.} DemoPSD
    \citep{demopsd2026} keeps the standard privileged information but decides
    \emph{how far} to follow the teacher at each token. The criterion is the
    \emph{disagreement} between teacher and student. When the two distributions
    are close, the teacher is followed. When they diverge too far, the teacher
    is ignored, a sign that the privileged information has influenced it too
    strongly. DemoPSD outperforms both GRPO and SDPO, and shows that
    performance and diversity can go together.
\end{itemize}
These methods shift the question, from which information to give towards where and how far to follow the teacher. Privileged information distorts the teacher's distribution on only a small fraction of tokens.

\paragraph{What remains open.}
\begin{itemize}
    \item \textbf{Crossing the \emph{kinds} of privileged information in a
    single protocol.} No work compares the oracle, the CoT, the plan, the
    rubric and feedback under strict self-distillation, on the same models and
    the same benchmarks. The two existing comparisons cover three forms under
    self-distillation \citep{kara2026alignment}, or five forms outside it
    \citep{dopd2026}.

    \item \textbf{Handling branching points.} \citet{kaur2026rethinking} call
    for explicitly preserving the places at which reasoning can diverge, so as
    to guide the student without crushing its deliberation.

    \item \textbf{Making the dosage adaptive.} The $\lambda$ of AR-OPD is
    fixed, and the probe of Purified OPSD has been applied to a single kind of
    information. Extending them to several kinds would reveal which carry the
    most transferable signal.
\end{itemize}

\medskip
\begin{axe}
\paragraph{\textcolor{accent}{Key takeaway --- Privileged information}} The question is no longer whether the teacher should receive privileged information, but which kind, how much of it, and how far it should be followed. Leakage is now named and measured \citep{aropd2026, demopsd2026}, and several remedies exist. The safety criterion: privileged information helps only if the student can reconstruct it at test time.
\\[2pt]
The final answer is the least transferable choice, since a teacher that already knows it expresses almost no uncertainty and encourages the student to take shortcuts. What transfers best is intermediate abstraction, meaning hints, plans and error-aligned critiques, which convey a skill without entailing the answer.
\end{axe}


\bigskip
\subsection{Lever C --- Loop Stability: Temporal Dynamics}
\label{sec:A4}

\bigskip
The preceding levers examined the distillation signal in itself. This one adds
time. In OPSD the teacher is the model frozen at its initial policy, so the
learning target stays put while the student moves away from it. The asymmetry
between them is twofold: temporal, between frozen weights and updated ones, and
informational, between privileged context and none.

\paragraph{Keeping the teacher useful without letting it collapse.} Two simple
answers fail. Freezing the teacher at its initial state quickly makes it
obsolete, since it ignores the student's progress. Updating it in step with the
student collapses the loop: the teacher ceases to be a reference external to
the student's trajectory and starts absorbing its drift. The loop confirms
itself instead of correcting itself.

The problem is not new. Self-supervised learning in vision solved it five years
ago, in an identical loop where a model learns from a copy of itself. Three
rules transfer directly to OPSD.
\begin{enumerate}
    \item Never backpropagate the gradient into the teacher. Its output is
    treated as a fixed target at every step. This is the \emph{stop-gradient}.
    SimSiam \citep{chen2021simsiam} shows that it is the decisive ingredient:
    without it, the model collapses to $0.1\,\%$ accuracy. OPSD already applies
    the rule, since no gradient flows back into the teacher, and goes further
    still, as the teacher's weights remain those of the initial policy.

    \item Let the teacher evolve, but more \textit{slowly} than the student.
    Rather than copying it from the student, it is built as an exponential
    moving average (EMA) of its past weights:
    \[
    \theta_T \;\leftarrow\; \rho\,\theta_T + (1-\rho)\,\theta_S
    \]
    The teacher ``lags behind'' the student, providing targets that are stable
    yet improving. Mean Teacher \citep{tarvainen2017meanteacher} is its origin,
    and \citet{busbridge2023ema} show that the momentum $\rho$ must be
    recalibrated when the batch size changes, at the risk of destabilizing the
    dynamics. This has never been tested directly on an LLM teacher.

    \item Maintain an \emph{asymmetry} between the two roles. BYOL
    \citep{grill2020byol} shows that a slow teacher is not enough: the student
    must also be structurally different from the teacher, or the loop converges
    to a constant. In OPSD this asymmetry is automatic, since the teacher sees
    the privileged information and the student does not. That imbalance is what
    prevents the two models from merging.
\end{enumerate}
The three rules are not mutually independent, and the vision literature is
itself divided on the second. SimSiam \citep{chen2021simsiam} shows that a
momentum encoder is \emph{not} necessary, the stop-gradient alone being enough
to prevent collapse, whereas BYOL and Mean Teacher rely on it. That
dissociation has never been tested in the LLM setting.

On the LLM side, two works from 2026 specify \emph{when} and \emph{by how much}
to move the teacher. CGTR \citep{cgtr2026} addresses timing. Refreshing the
teacher at a fixed interval can lock it onto a student that is drifting, which
destabilizes the whole loop; they call this failure \emph{state-oblivious
collapse}. The remedy is to refresh only once the student has genuinely
progressed, as measured by a reward gain. What buys stability, they argue, is
the \emph{isolation period} between two refreshes, during which the teacher is
entirely frozen and does not absorb the student's drift. On that argument, an
EMA teacher has no isolation period at all, since it absorbs a fraction of the
drift at every step. TOP-D \citep{topd2026} addresses distance. A teacher too
far from the student produces large, noisy gradients, and training diverges.
They therefore keep the teacher \emph{close}, in the manner of a trust region,
which bounds gradient variance and guarantees steady improvement.

\paragraph{Does the student forget?} The second risk is \emph{catastrophic
forgetting}. In training on a new task, a model moves its weights. That
displacement improves the new task but can erase what the weights previously
encoded: the model gains one capability and loses another. The whole question
is therefore how far the model moves.

RL's Razor \citep{shenfeld2025rlrazor} gives the sharpest answer. Forgetting,
they show, is predicted by a single quantity: the KL distance between the
trained policy and the starting policy, measured on the new task. Forgetting
therefore grows with displacement. And on-policy RL, among all the ways of
solving the task, selects the one that moves the model least. It thus forgets
less than SFT, which can travel arbitrarily far. Since OPSD is on-policy, it
should inherit this caution. Two results, however, appear to conflict:
\begin{itemize}
    \item SDFT \citep{shenfeld2026sdft} shows that self-distillation
    \emph{reduces} forgetting: having the model produce its own version of the
    answer keeps it close to itself, so the weights change little.

    \item Denser $\neq$ Better \citep{wang2026denser} shows the opposite: in
    continual learning, \emph{dense} self-distillation forgets \emph{more} than
    sparse RL, and can even collapse.
\end{itemize}
We suggest that the conflict is only apparent. Self-distillation (SDFT) changes
the weights less than SFT, but more than sparse RL, since a token-level signal
constrains the model far more than a single terminal reward. The ordering this
suggests is:
\[
    \text{forgetting(raw SFT)} \;>\; \text{forgetting(dense self-distillation)}
    \;>\; \text{forgetting(sparse RL)} .
\]
The conjecture is itself contested. \citet{hubotter2026sdpo} evaluate their
final checkpoints on held-out tasks and report that SDPO, a dense
self-distillation method, degrades prior capabilities \emph{less} than GRPO,
which would reverse the last two terms. The protocols differ, a single task
against a sequence of tasks, but the discrepancy is unresolved.

The real question is then how to move OPSD towards the favourable end,
acquiring capabilities without losing the old ones. Two methods attempt this:
\begin{itemize}
    \item MOPD \citep{mopd2026} trains one specialist teacher per domain, then
    distils them all into a single student on that student's own rollouts. The
    approach is already deployed in an industrial model.

    \item CaMOPD \citep{camopd2026} starts from a finer observation: the
    gradient that \emph{recovers} general capabilities and the one that
    \emph{preserves} the domain often point in opposite directions and cancel
    out. Their solution is to apply them in alternation rather than to add
    them.
\end{itemize}

\paragraph{Scheduling guidance over time.} A last lever lets guidance
\emph{decay} as the student progresses. We will call this the \textbf{decay of
privileged information}, to distinguish it from the update of the teacher's
weights discussed above. These are two independent mechanisms that a loose
vocabulary often conflates. If the privileged information is the full chain of
thought, the teacher can be shown all of it at the outset, when the student is
still weak, and a growing share can then be withheld until the student reasons
on its own.

This idea is well studied in a neighbouring setting: off-policy guidance, where
the help is a solution prefix rather than a conditioning of the teacher. R3
\citep{xi2024r3} is the prototype. The student is given the problem $x$ along
with a large part of the solution $y$, so that only a few tokens remain to be
produced; that starting point is then pushed back until the student generates
everything alone. Prefix-RFT \citep{huang2026prefixrft} takes up the idea: a
prefix of the demonstration is given, and its length reduced over the course of
training. AdaBack \citep{amani2026adaback} makes it \emph{adaptive}: the share
revealed is no longer fixed in advance but adjusted example by example
according to the student's success. A problem still failed receives more help,
a mastered one receives less. UFT \citep{liu2025uft} supplies the theoretical
justification: without initial help, a weak model takes exponentially long to
stumble upon a good trajectory. Decaying guidance is therefore not a
convenience but a condition of convergence.

All of this holds for off-policy guidance. The transposition to OPSD was begun
by ATESD \citep{atesd2026}. In standard OPSD, the teacher always sees the
reference trace in full; they call this defect the \emph{teacher-side exposure
mismatch}. A sweep at fixed exposure shows two things: (i) full exposure is not
reliably the best choice; and (ii) the disagreement between teacher and student
grows monotonically with the amount of reasoning revealed. This second point
connects directly to the PMI mechanism of \S\ref{sec:A2}. Rather than fixing in
advance the share of the trace shown to the teacher, they let it vary during
training. A small controller adjusts that share automatically, driven by the
rate at which the student is progressing. When the student learns quickly, less
can be shown; when it stalls, more is revealed. The mechanism remains embedded
in the dense per-token loss, with no separate stage. Gains range from $+0.95$
to $+2.33$ in avg@12 on Qwen3-1.7B/4B/8B. PAINT \citep{paint2026} follows a
related logic, through adaptive masking of the verified solution.

The principle is thus established, and what remains open concerns mainly the
\textit{how}. A deterministic schedule rather than a learned controller is one
option. ATESD varies the \textit{quantity} of information revealed; an avenue
still unexplored is to vary its \textit{nature}, moving from the oracle to a
plan and then to a hint as training proceeds.

\paragraph{What remains open.}
\begin{itemize}
    \item \textbf{Comparing teacher update rules at fixed privileged
    information}: frozen teacher, EMA, gated refresh, proximal teacher. The
    only existing comparison is an ablation internal to a single paper
    \citep{hubotter2026sdpo}, on one kind of privileged information and at a
    single momentum value. The systematic comparison \citep{cgtr2026} is
    conducted without any privileged information at all, with the teacher
    reduced to a past checkpoint.

    \item \textbf{Placing OPSD on the forgetting scale} (SFT $>$ dense
    self-distillation $>$ sparse RL), and finding how to move it towards the
    end that best preserves the base model's knowledge while still acquiring
    new capabilities.

    \item \textbf{Transposing the tools of vision.} This corpus remains largely
    untapped on the LLM side. The calibration of EMA momentum
    \citep{busbridge2023ema}, for instance, has never been tested on an LLM
    teacher, even though collapse is more aggressive at small scale.
\end{itemize}

\bigskip
\begin{axe}
\paragraph{\textcolor{accent}{Key takeaway --- Loop stability}} The teacher is the model frozen at its initial policy, so the target stays put while the student moves away from it. Freezing the weights imposes a performance ceiling, while updating them at the student's own pace removes all anchoring. That ceiling comes from the freezing of the weights and not from the stop-gradient, which remains necessary at each step. Two distinct temporal
levers follow: letting the teacher's weights evolve, and letting the information given to it decay.
\end{axe}


\bigskip
\section{Synthesis: Three Levers, One Symptom}

\medskip
The initial bet of OPSD was that density was the decisive variable: with a
token-level signal, learning would be both more efficient and more effective.
Six months on, that bet has shifted. Density itself is no longer the variable
that matters. What matters is where the signal is applied, what it encodes, and
when it is allowed to change. The teacher's signal cuts both ways: it can
convey skills to the student, and it can equally teach it shortcuts unavailable
at inference time. This is why controlling the signal matters more than
increasing it.

\begin{figure}[!t]
  \centering
  \includegraphics[width=\textwidth]{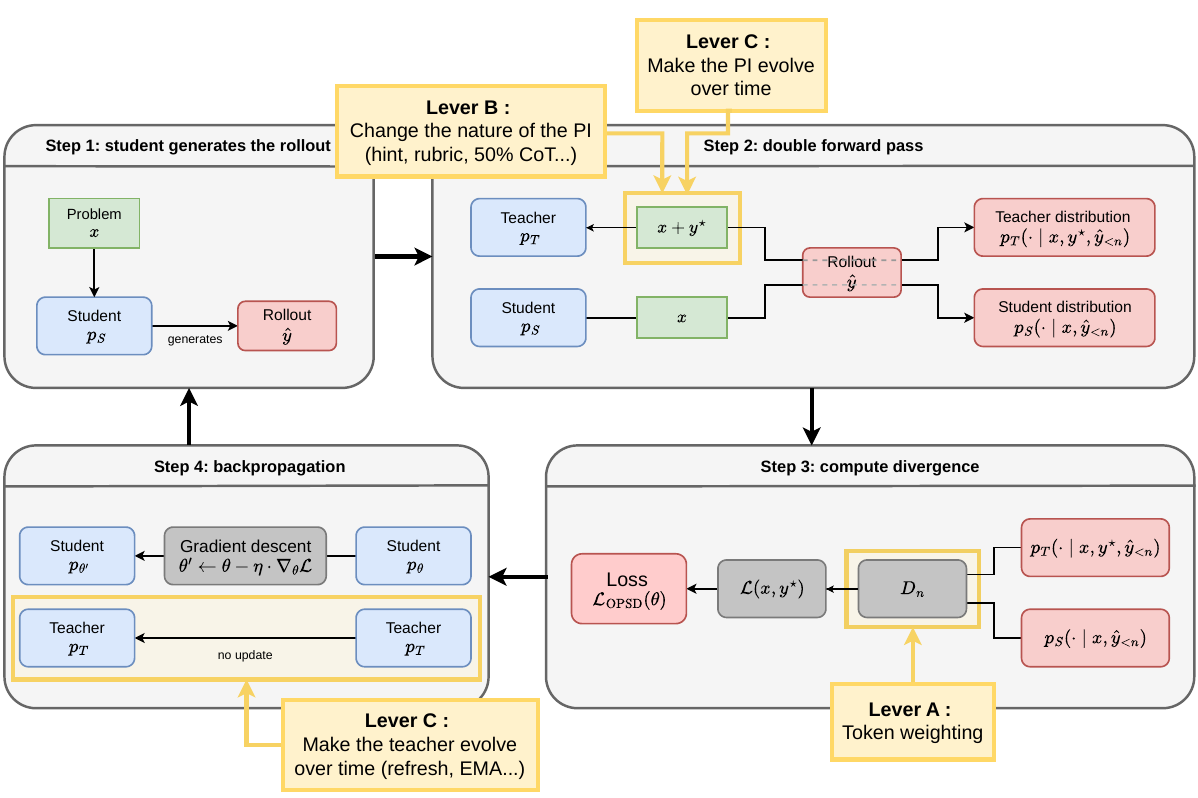}
  \caption{The three levers, mapped onto the loop of
  Figure~\ref{fig:schema-OPSD}. Lever C appears at two points: the teacher's
  exposure to the privileged information and the update of its weights are two
  independent mechanisms.}
  \label{fig:leviers}
\end{figure}

Collapse is not one of the levers. It is the consequence of a poorly
transmitted signal, and therefore the measure against which the three are
judged, read through pass@$k$ rather than through the mean score or the entropy
(\S\ref{sec:A2}). Figure~\ref{fig:leviers} locates each lever on the training
loop.

\paragraph{Axis A. Where? Weighting the tokens.}

In the founding paper \citep{zhao2026opsd}, all tokens are weighted uniformly.
This is the simplest choice, and it has proved suboptimal. A rollout of at most
$1{,}024$ tokens contains only a small number of decisions that genuinely
commit the reasoning; the rest is formatting. Weighting uniformly therefore
dilutes the signal attached to reasoning, and lets formatting occupy most of
the gradient. Since then, several teams have shown that a small fraction of
tokens carries most of the learning signal \citep{armandpour2026unmasking}, and
a number of papers have proposed ways of making that signal selective, notably
by exploiting entropy \citep{entropyaware2026}. The objective is always the
same: to shield the student from noise and from collapse.

No method has yet reached consensus, however. The central trade-off runs between a gain in accuracy and a loss in diversity \citep{li2025dphrl}.

\paragraph{Axis B. What? Choosing the privileged information.}

In the founding paper, the privileged information is the reference solution \citep{zhao2026opsd}. That is the simplest thing to hand the teacher, and it turns out to be the wrong one. The intuition that ``the better informed the teacher, the better the signal'' has since been refuted \citep{kaur2026rethinking}. A teacher that already knows the final answer no longer needs to deliberate or to reason. It ceases to express uncertainty \citep{kim2026why}, which pushes the student to skip steps and to behave as though it already knew the answer. The student thereby becomes overconfident and loses substantial diversity \citep{nicolicioiu2026} and out-of-domain capability \citep{kim2026why}.

The search since then has been for a kind of privileged information that conveys a skill without entailing the answer. Several papers report gains in
both performance and diversity using hints and plans \citep{dopd2026}, error-aligned critiques \citep{kara2026alignment}, or environment feedback
\citep{hubotter2026sdpo}. What separates them from the reference solution is the reconstruction criterion of \S\ref{sec:A3}.

Even so, no kind of privileged information has reached consensus. One avenue remains little explored: varying both the nature and the quantity of what the teacher is shown, according to context.

\paragraph{Axis C. When? Making the teacher dynamic.}

In the founding paper, the teacher is frozen at the initial policy. The
consequence is immediate: as the student progresses, its target stays where it
was and the gain diminishes. This ceiling comes from the freezing of the
weights, and not from the stop-gradient, which prevents the teacher from
sliding towards the student within a step and remains necessary
\citep{chen2021simsiam}. The field is therefore asking how to refresh the
weights without collapsing the model.

Several methods exist. The teacher can be made to evolve as a moving average of
the student's weights \citep{tarvainen2017meanteacher}, or refreshed only once
the student's performance has improved \citep{cgtr2026}. A second, independent
lever is to let the privileged information itself decay as training proceeds
\citep{atesd2026}. This axis nonetheless remains the least explored of the
three, and the difficulty is as much methodological as conceptual: the
collapses documented in this family occur after several hundred steps
\citep{cgtr2026}, and a short run does not see them.

\bigskip
\section*{Conclusion}

OPSD proposed to replace the capability advantage of an external teacher with
an information asymmetry between a model and itself. Judged first on accuracy,
the method appeared effective and more frugal in generated tokens. Six months
later the picture is more nuanced, because the asymmetry that produces the
signal is also the one that biases it.

\bigskip
Three results are established today. Privileged information helps only if the
student can reconstruct it on its own at test time, and harms it as soon as it
must be presupposed. It follows that the reference solution, the most natural
choice, is among those tested the least transferable. And collapse appears
neither in the mean score nor in the entropy: a self-distilled model can
display a higher entropy than an RL-trained one while producing fewer distinct
lines of reasoning, and only pass@$k$ reveals it. These results were already
available, but scattered and named differently from one paper to the next. Our
contribution is to unify them, and to reconnect them to the theory of learning
using privileged information, from which the field has borrowed the vocabulary
without the results.

\bigskip
Should OPSD then be used in production today? No, not naively. It is a
research technique whose failure modes are by now well documented: on reasoning
models, poorly chosen privileged information degrades performance instead of
improving it \citep{kaur2026rethinking}; in continual learning, the dense
version forgets more than standard RL and can collapse \citep{wang2026denser}.
It remains a promising direction under two conditions. The first is to use it
with the safeguards set out in \S\ref{sec:eval}. The second is to see it for
what it is today, a post-SFT fine-tuning method that is frugal in generated
tokens and requires no external teacher, rather than a turnkey training method.
For an organization seeking to control the production of its own small models,
the method is genuinely appealing, but it still lies at the research stage.

\bigskip
Our analysis has several limitations. We have treated only mathematical
reasoning, leaving out the multimodal and agentic branches. Nearly all the work
cited consists of preprints less than six months old, conducted for the most
part on a single model family and at sizes not exceeding a few billion
parameters, a ceiling the founding paper itself acknowledges, citing compute
constraints. The conclusions above should be read with that reservation in
mind.

\bigskip
A model can guide itself, provided that the teaching side holds information the
answering side does not. Six months of work have qualified that principle
rather than overturned it. Everything turns on how the asymmetry is controlled.

\newpage
\bibliographystyle{plainnat}
\bibliography{references}

@article{zhao2026opsd,
  title={Self-Distilled Reasoner: {On-Policy} {Self-Distillation} for Large Language Models},
  author={Zhao, Siyan and Xie, Zhihui and Liu, Mengchen and Huang, Jing and Pang, Guan and Chen, Feiyu and Grover, Aditya},
  journal={arXiv preprint arXiv:2601.18734},
  note={{ICLM} 2026},
  year={2026}
}

@article{hubotter2026sdpo,
  title   = {Reinforcement Learning via Self-Distillation},
  author  = {H{\"u}botter, Jonas and L{\"u}beck, Frederike and Behric, Lejs and Baumann, Anton and Bagatella, Marco and Marta, Daniel and Hakimi, Ido and Shenfeld, Idan and Kleine B{\"u}ning, Thomas and Guestrin, Carlos and Krause, Andreas},
  journal = {arXiv preprint arXiv:2601.20802},
  year    = {2026},
  url     = {https://arxiv.org/abs/2601.20802}
}

@article{shenfeld2026sdft,
  title={Self-Distillation Enables Continual Learning},
  author={Shenfeld, Idan and Damani, Mehul and H{\"u}botter, Jonas and Agrawal, Pulkit},
  journal={arXiv preprint arXiv:2601.19897},
  note={ICML 2026 (poster)},
  year={2026}
}

@article{kim2026why,
  title={Why Does Self-Distillation (Sometimes) Degrade the Reasoning Capability of {LLMs}?},
  author={Kim, Jeonghye and Luo, Xufang and Kim, Minbeom and Lee, Sangmook and Kim, Dohyung and Jeon, Jiwon and Li, Dongsheng and Yang, Yuqing},
  journal={arXiv preprint arXiv:2603.24472},
  year={2026}
}

@article{kaur2026rethinking,
  title={Rethinking On-Policy Self-Distillation for Thinking Models},
  author={Kaur, Simran and Ri, Narutatsu and He, Yinghui and Fowl, Liam and Arora, Sanjeev},
  journal={arXiv preprint arXiv:2607.05184},
  year={2026}
}

@article{demopsd2026,
  title={{DemoPSD}: Disagreement-Modulated Policy Self-Distillation},
  author={Li, Yunhe and Shi, Hao and Liu, Wenhao and Ruan, Mengzhe and Hou, Hanxu and Dai, Zhongxiang and Qiu, Shuang and Song, Linqi},
  journal={arXiv preprint arXiv:2607.02502},
  year={2026}
}

@article{wang2026denser,
  title={Denser {$\neq$} Better: Limits of On-Policy Self-Distillation for Continual Post-Training},
  author={Wang, Meng and Zhao, Haohan and Liu, Wenzhuo and Yang, Lu and Liu, Geng and Guo, Haiyang and Xie, Guo-Sen and Meng, Gaofeng and Liu, Hongbin and Zhu, Fei},
  journal={arXiv preprint arXiv:2607.01763},
  year={2026}
}

@article{purifiedopsd2026,
  title={Purified {OPSD}: On-Policy Self-Distillation Without Losing How to Think},
  author={Shen, Zhanming and Tong, Jintao and Yan, Shaotian and Shen, Chen and Chen, Hao and Ye, Wentao and Hu, Xiaomeng and Miao, Rui and Wang, Haobo and Zhao, Junbo and Chen, Gang and Ye, Jieping},
  journal={arXiv preprint arXiv:2607.02234},
  year={2026}
}

@article{aropd2026,
  title={Beyond Absolute Imitation: Anchored Residual Guidance for Privileged On-Policy Distillation},
  author={Zhang, Wenhao},
  journal={arXiv preprint arXiv:2606.10385},
  year={2026}
}

@article{nicolicioiu2026,
  title={On-Policy Self-Distillation with Sampled Demonstrations Reduces Output Diversity},
  author={Nicolicioiu, Andrei Liviu and Pezeshki, Mohammad and Courville, Aaron},
  journal={arXiv preprint arXiv:2606.26091},
  year={2026}
}

@article{dopd2026,
  title   = {{DOPD}: Dual On-policy Distillation},
  author  = {Yu, Xinlei and Li, Gen and Si, Qingyi and Zhang, Guibin and Xu, Yuqi and Wang, Congcong and Dong, Shuai and Tuo, Kaiwen and Zeng, Xiangyu and Feng, Kaituo and Wang, Qunzhong and Shi, Yang and Hu, Xiaobin and Yue, Xiangyu and Wang, Jiaqi and Yan, Shuicheng},
  journal = {arXiv preprint arXiv:2606.30626},
  year    = {2026},
  url     = {https://arxiv.org/abs/2606.30626}
}

@article{kara2026alignment,
  title   = {The Role of Feedback Alignment in Self-Distillation},
  author  = {Kara, Semih and Ersoy, O{\u{g}}uzhan},
  journal = {arXiv preprint arXiv:2606.11173},
  year    = {2026},
  url     = {https://arxiv.org/abs/2606.11173}
}

@article{antisd2026pmi,
  title={Anti-Self-Distillation for Reasoning {RL} via Pointwise Mutual Information},
  author={Shen, Guobin and Cheng, Xiang and Zhao, Chenxiao and Huang, Lei and Li, Jindong and Zhao, Dongcheng and Yu, Xing},
  journal={arXiv preprint arXiv:2605.11609},
  year={2026}
}

@article{rubric2026,
  title={Rethinking Reward Supervision: Rubric-Conditioned Self-Distillation},
  author={Gu, Siyi and Chen, Jialin and Zhou, Sophia and Cohan, Arman and Ying, Rex},
  journal={arXiv preprint arXiv:2606.19327},
  year={2026}
}

@article{mopd2026,
  title={{MOPD}: Multi-Teacher On-Policy Distillation for Capability Integration in {LLM} Post-Training},
  author={Ma, Wenhan and Wei, Jianyu and Zhao, Liang and Zhang, Hailin and Xiao, Bangjun and Li, Lei and Yang, Qibin and Gao, Bofei and Wang, Yudong and Li, Rang and Dong, Jinhao and Sui, Zhifang and Luo, Fuli},
  journal={arXiv preprint arXiv:2606.30406},
  year={2026}
}

@article{cgtr2026,
  title={When Should the Teacher Move? Temporal Coupling and Stability in Self On-Policy Distillation},
  author={Guo, Haowei and Bi, Baolong and Zhang, Ruicheng and Sun, Bingqian and Zhang, Wentao},
  journal={arXiv preprint arXiv:2606.03532},
  year={2026}
}

@article{paint2026,
  title   = {{PAINT}: Partial-Solution Adaptive Interpolated Training for Self-Distilled Reasoners},
  author  = {Tan, Zhiquan and Hong, Yinrong},
  journal = {arXiv preprint arXiv:2604.26573},
  year    = {2026},
  url     = {https://arxiv.org/abs/2604.26573}
}

@article{atesd2026,
  title   = {Adaptive Teacher Exposure for Self-Distillation in {LLM} Reasoning},
  author  = {Han, Zihao and Zhang, Tiangang and Wang, Huaibin and Sun, Yilun},
  journal = {arXiv preprint arXiv:2605.11458v3},
  year    = {2026},
  url     = {https://arxiv.org/abs/2605.11458}
}

@article{topd2026,
  title={Trust Region Policy Distillation},
  author={Xie, Zhengpeng and Zhang, Li Lyna and Xie, Zeke and Yang, Mao},
  journal={arXiv preprint arXiv:2607.04751},
  year={2026}
}

@article{entropyaware2026,
  title={Entropy-Aware On-Policy Distillation of Language Models},
  author={Jin, Woogyeol and Min, Taywon and Yang, Yongjin and Wei, Dennis and Zhou, Yi and Ravindra Kadhe, Swanand and Baracaldo, Nathalie and Lee, Kimin},
  journal={arXiv preprint arXiv:2603.07079},
  note={ICML 2026},
  year={2026}
}

@article{camopd2026,
  title={Counteraction-Aware Multi-Teacher On-Policy Distillation for General Capability Recovery with Domain Preservation},
  author={Chen, Tianlei and Ou, Jiao and Liu, Ziyuan and Tang, Ruiming and Liang, Jian and Li, Han},
  journal={arXiv preprint arXiv:2605.27115},
  year={2026}
}

@inproceedings{agarwal2024gkd,
  title     = {{On-Policy} Distillation of Language Models: Learning from Self-Generated Mistakes},
  author    = {Agarwal, Rishabh and Vieillard, Nino and Zhou, Yongchao and Stanczyk, Piotr and Ramos, Sabela and Geist, Matthieu and Bachem, Olivier},
  booktitle = {International Conference on Learning Representations (ICLR)},
  year      = {2024},
  url       = {https://arxiv.org/abs/2306.13649}
}

@inproceedings{gu2024minillm,
  title={Mini{LLM}: Knowledge Distillation of Large Language Models},
  author={Gu, Yuxian and Dong, Li and Wei, Furu and Huang, Minlie},
  booktitle={International Conference on Learning Representations (ICLR)},
  note={arXiv:2306.08543},
  year={2024}
}

@inproceedings{ko2024distillm,
  title={Disti{LLM}: Towards Streamlined Distillation for Large Language Models},
  author={Ko, Jongwoo and Kim, Sungnyun and Chen, Tianyi and Yun, Se-Young},
  booktitle={International Conference on Machine Learning (ICML)},
  note={arXiv:2402.03898},
  year={2024}
}

@misc{lu2025blog,
  title = {On-Policy Distillation},
  author = {Lu, Kevin},
  howpublished = {Thinking Machines Lab, Connectionism},
  year = {2025},
  month = oct,
  doi = {10.64434/tml.20251026},
  url = {https://thinkingmachines.ai/blog/on-policy-distillation/},
  note = {Accessed: 2026-08-07}
}

@article{shao2024grpo,
  title   = {{DeepSeekMath}: Pushing the Limits of Mathematical Reasoning in Open Language Models},
  author  = {Shao, Zhihong and Wang, Peiyi and Zhu, Qihao and Xu, Runxin and Song, Junxiao and Bi, Xiao and Zhang, Haowei and Zhang, Mingchuan and Li, Y. K. and Wu, Y. and Guo, Daya},
  journal = {arXiv preprint arXiv:2402.03300},
  year    = {2024},
  url     = {https://arxiv.org/abs/2402.03300}
}

@article{vapnik2009lupi,
  title   = {A New Learning Paradigm: Learning Using Privileged Information},
  author  = {Vapnik, Vladimir and Vashist, Akshay},
  journal = {Neural Networks},
  volume  = {22},
  number  = {5--6},
  pages   = {544--557},
  year    = {2009},
  doi     = {10.1016/j.neunet.2009.06.042}
}

@inproceedings{lopezpaz2016unifying,
  title={Unifying Distillation and Privileged Information},
  author={Lopez-Paz, David and Bottou, L{\'e}on and Sch{\"o}lkopf, Bernhard and Vapnik, Vladimir},
  booktitle={International Conference on Learning Representations (ICLR)},
  note={arXiv:1511.03643},
  year={2016}
}

@inproceedings{weihs2021advisor,
  title     = {Bridging the Imitation Gap by Adaptive Insubordination},
  author    = {Weihs, Luca and Jain, Unnat and Liu, Iou-Jen and
               Salvador, Jordi and Lazebnik, Svetlana and Kembhavi, Aniruddha
               and Schwing, Alexander},
  booktitle = {Advances in Neural Information Processing Systems (NeurIPS)},
  year      = {2021},
  url       = {https://arxiv.org/abs/2007.12173}
}

@inproceedings{chen2019cheating,
  title={Learning by Cheating},
  author={Chen, Dian and Zhou, Brady and Koltun, Vladlen and Kr{\"a}henb{\"u}hl, Philipp},
  booktitle={Conference on Robot Learning (CoRL)},
  year={2019}
}

@inproceedings{kumar2021rma,
  title={{RMA}: Rapid Motor Adaptation for Legged Robots},
  author={Kumar, Ashish and Fu, Zipeng and Pathak, Deepak and Malik, Jitendra},
  booktitle={Robotics: Science and Systems (RSS)},
  year={2021}
}

@inproceedings{furlanello2018ban,
  title     = {Born-Again Neural Networks},
  author    = {Furlanello, Tommaso and Lipton, Zachary C. and Tschannen, Michael and Itti, Laurent and Anandkumar, Anima},
  booktitle = {International Conference on Machine Learning (ICML)},
  year      = {2018},
  url       = {https://arxiv.org/abs/1805.04770}
}

@inproceedings{mobahi2020self,
  title     = {Self-Distillation Amplifies Regularization in {Hilbert} Space},
  author    = {Mobahi, Hossein and Farajtabar, Mehrdad and Bartlett, Peter L.},
  booktitle = {Advances in Neural Information Processing Systems (NeurIPS)},
  year      = {2020},
  url       = {https://arxiv.org/abs/2002.05715}
}

@inproceedings{zelikman2022star,
  title     = {{STaR}: Bootstrapping Reasoning With Reasoning},
  author    = {Zelikman, Eric and Wu, Yuhuai and Mu, Jesse and Goodman, Noah D.},
  booktitle = {Advances in Neural Information Processing Systems (NeurIPS)},
  year      = {2022},
  url       = {https://arxiv.org/abs/2203.14465}
}

@article{shumailov2024nature,
  title={{AI} Models Collapse When Trained on Recursively Generated Data},
  author={Shumailov, Ilia and Shumaylov, Zakhar and Zhao, Yiren and Papernot, Nicolas and Anderson, Ross and Gal, Yarin},
  journal={Nature},
  volume={631},
  pages={755--759},
  year={2024},
  dio= {10.1038/s41586-024-07566-y}
}

@article{gerstgrasser2024accumulate,
  title   = {Is Model Collapse Inevitable? Breaking the Curse of Recursion by Accumulating Real and Synthetic Data},
  author  = {Gerstgrasser, Matthias and Schaeffer, Rylan and Dey, Apratim and Rafailov, Rafael and Sleight, Henry and Hughes, John and Korbak, Tomasz and Agrawal, Rajashree and Pai, Dhruv and Gromov, Andrey and Roberts, Daniel A. and Yang, Diyi and Donoho, David L. and Koyejo, Sanmi},
  journal = {arXiv preprint arXiv:2404.01413},
  year    = {2024},
  url     = {https://arxiv.org/abs/2404.01413}
}

@inproceedings{yue2025rl,
  title={Does Reinforcement Learning Really Incentivize Reasoning Capacity in {LLMs} Beyond the Base Model?},
  author={Yue, Yang and Chen, Zhiqi and Lu, Rui and Zhao, Andrew and Wang, Zhaokai and Song, Shiji and Huang, Gao},
  booktitle={Advances in Neural Information Processing Systems (NeurIPS), Oral},
  note={arXiv:2504.13837},
  year={2025}
}

@article{li2025dphrl,
  title={The Choice of Divergence: A Neglected Key to Mitigating Diversity Collapse in Reinforcement Learning with Verifiable Reward},
  author={Li, Long and Zhou, Zhijian and Hao, Jiaran and Klein Liu, Jason and Miao, Yanting and Pang, Wei and Tan, Xiaoyu and Chu, Wei and Wang, Zhe and Pan, Shirui and Qu, Chao and Qi, Yuan},
  journal={arXiv preprint arXiv:2509.07430},
  year={2025}
}

@article{cui2025entropy,
  title={The Entropy Mechanism of Reinforcement Learning for Reasoning Language Models},
  author={Cui, Ganqu and Zhang, Yuchen and Chen, Jiacheng and Yuan, Lifan and Wang, Zhi and Zuo, Yuxin and Li, Haozhan and Fan, Yuchen and Chen, Huayu and Chen, Weize and Liu, Zhiyuan and Peng, Hao and Bai, Lei and Ouyang, Wanli and Cheng, Yu and Zhou, Bowen and Ding, Ning},
  journal={arXiv preprint arXiv:2505.22617},
  year={2025}
}

@article{dapo2025,
  title={{DAPO}: An Open-Source {LLM} Reinforcement Learning System at Scale},
  author={Yu, Qiying and Zhang, Zheng and Zhu, Ruofei and Yuan, Yufeng and Zuo, Xiaochen and Yue, Yu and Dai, Weinan and Fan, Tiantian and Liu, Gaohong and Liu, Lingjun and Liu, Xin and Lin, Haibin and Lin, Zhiqi and Ma, Bole and Sheng, Guangming and Tong, Yuxuan and Zhang, Chi and Zhang, Mofan and Zhang, Wang and Zhu, Hang and Zhu, Jinhua and Chen, Jiaze and Chen, Jiangjie and Wang, Chengyi and Yu, Hongli and Song, Yuxuan and Wei, Xiangpeng and Zhou, Hao and Liu, Jingjing and Ma, Wei-Ying and Zhang, Ya-Qin and Yan, Lin and Qiao, Mu and Wu, Yonghui and Wang, Mingxuan},
  journal={arXiv preprint arXiv:2503.14476},
  year={2025}
}

@inproceedings{shenfeld2025rlrazor,
  title={{RL}'s Razor: Why Online Reinforcement Learning Forgets Less},
  author={Shenfeld, Idan and Pari, Jyothish and Agrawal, Pulkit},
  booktitle={Advances in Neural Information Processing Systems (NeurIPS)},
  note={arXiv:2509.04259},
  year={2025}
}

@article{deepseekr1,
  title   = {{DeepSeek-R1} incentivizes reasoning in {LLMs} through reinforcement learning},
  author  = {{DeepSeek-AI}},
  journal = {Nature},
  volume  = {645},
  pages   = {633--638},
  year    = {2025},
  doi     = {10.1038/s41586-025-09422-z},
  note    = {Preprint: arXiv:2501.12948}
}

@article{armandpour2026unmasking,
  title={Unmasking On-Policy Distillation: Where It Helps, Where It Hurts, and Why},
  author={Armandpour, Mohammadreza and Ilhan, Fatih and Harrison, David and Jaiswal, Ajay and N.M Hoang, Duc and Faghri, Fartash and Zhang, Yizhe and Cho, Minsik and Farajtabar, Mehrdad},
  journal={arXiv preprint arXiv:2605.10889},
  year={2026}
}

@inproceedings{huang2026prefixrft,
  title={Blending Supervised and Reinforcement Fine-Tuning with Prefix Sampling},
  author={Huang, Zeyu and Cheng, Tianhao and Qiu, Zihan and Wang, Zili and Xu, Yinghui and Ponti, Edoardo M. and Titov, Ivan},
  booktitle={International Conference on Machine Learning (ICML)},
  note={arXiv:2507.01679},
  year={2026}
}

@article{amani2026adaback,
  title={Reinforcement Learning for Reasoning by Adaptively Revealing Rationales},
  author={Amani, Mohammad Hossein and Lotfi, Aryo and Baldwin, Nicolas M. and Bengio, Samy and Farajtabar, Mehrdad and Abbe, Emmanuel and West, Robert},
  journal={arXiv preprint arXiv:2506.18110},
  year={2025}
}

@article{liu2025uft,
  title={{UFT}: Unifying Supervised and Reinforcement Fine-Tuning},
  author={Liu, Mingyang and Farina, Gabriele and Ozdaglar, Asuman},
  journal={arXiv preprint arXiv:2505.16984},
  year={2025}
}

@inproceedings{xi2024r3,
  title={Training Large Language Models for Reasoning through Reverse Curriculum Reinforcement Learning},
  author={Xi, Zhiheng and Chen, Wenxiang and Hong, Boyang and Jin, Senjie and Zheng, Rui and He, Wei and Ding, Yiwen and Liu, Shichun and Guo, Xin and Wang, Junzhe and Guo, Honglin and Shen, Wei and Fan, Xiaoran and Zhou, Yuhao and Dou, Shihan and Wang, Xiao and Zhang, Xinbo and Sun, Peng and Gui, Tao and Zhang, Qi and Huang, Xuanjing},
  booktitle={International Conference on Machine Learning (ICML)},
  note={arXiv:2402.05808},
  year={2024}
}

@inproceedings{tarvainen2017meanteacher,
  title={Mean Teachers are Better Role Models: Weight-Averaged Consistency Targets improve semi-supervised deep learning results},
  author={Tarvainen, Antti and Valpola, Harri},
  booktitle={Advances in Neural Information Processing Systems (NeurIPS)},
  year={2017}
}

@inproceedings{grill2020byol,
  title={Bootstrap Your Own Latent: A New Approach to Self-Supervised Learning},
  author={Grill, Jean-Bastien and Strub, Florian and Altch{\'e}, Florent and Tallec, Corentin and Richemond, Pierre H. and Buchatskaya, Elena and Doersch, Carl and Avila Pires, Bernardo and Guo, Zhaohan Daniel and Gheshlaghi Azar, Mohammad and Piot, Bilal and Kavukcuoglu, Koray and Munos, R{\'e}mi and Valko, Michal},
  booktitle={Advances in Neural Information Processing Systems (NeurIPS)},
  note={arXiv:2006.07733},
  year={2020}
}

@inproceedings{chen2021simsiam,
  title={Exploring Simple Siamese Representation Learning},
  author={Chen, Xinlei and He, Kaiming},
  booktitle={IEEE/CVF Conference on Computer Vision and Pattern Recognition (CVPR)},
  note={arXiv:2011.10566},
  year={2021}
}

@inproceedings{busbridge2023ema,
  title={How to Scale Your {EMA}},
  author={Busbridge, Dan and Ramapuram, Jason and Ablin, Pierre and Likhomanenko, Tatiana and Dhekane, Eeshan Gunesh and Suau, Xavier and Webb, Russ},
  booktitle={Advances in Neural Information Processing Systems (NeurIPS)},
  note={arXiv:2307.13813},
  year={2023}
}

@inproceedings{hochlehnert2025sober,
  title={A Sober Look at Progress in Language Model Reasoning: Pitfalls and Paths to Reproducibility},
  author={Hochlehnert, Andreas and Bhatnagar, Hardik and Udandarao, Vishaal and Albanie, Samuel and Prabhu, Ameya and Bethge, Matthias},
  booktitle={Conference on Language Modeling (COLM)},
  note={arXiv:2504.07086},
  year={2025}
}

@article{shao2025spurious,
  title={Spurious Rewards: Rethinking Training Signals in {RLVR}},
  author={Shao, Rulin and Li, Shuyue Stella and Xin, Rui and Geng, Scott and Wang, Yiping and Oh, Sewoong and Du, Simon S. and Lambert, Nathan and Min, Sewon and Krishna, Ranjay and Tsvetkov, Yulia and Hajishirzi, Hannaneh and Koh, Pang Wei and Zettlemoyer, Luke},
  journal={arXiv preprint arXiv:2506.10947},
  year={2025}
}

@inproceedings{wu2026reasoning,
  title={Reasoning or Memorization? Unreliable Results of Reinforcement Learning Due to Data Contamination},
  author={Wu, Mingqi and Zhang, Zhihao and Dong, Qiaole and Xi, Zhiheng and Zhao, Jun and Jin, Senjie and Fan, Xiaoran and Zhou, Yuhao and Lv, Huijie and Zhang, Ming and Fu, Yanwei and Liu, Qin and Zhang, Songyang and Zhang, Qi},
  booktitle={AAAI Conference on Artificial Intelligence},
  note={arXiv:2507.10532},
  year={2026}
}

@inproceedings{liu2025gpassk,
  title={Are Your {LLMs} Capable of Stable Reasoning?},
  author={Liu, Junnan and Liu, Hongwei and Xiao, Linchen and Wang, Ziyi and Liu, Kuikun and Gao, Songyang and Zhang, Wenwei and Zhang, Songyang and Chen, Kai},
  booktitle={Findings of the ACL},
  note={arXiv:2412.13147},
  year={2025}
}

@article{matharena2025,
  title={{MathArena}: Evaluating {LLMs} on Uncontaminated Math Competitions},
  author={Balunovi{\'c}, Mislav and Dekoninck, Jasper and Petrov, Ivo and Jovanovi{\'c}, Nikola and Vechev, Martin},
  journal={arXiv preprint arXiv:2505.23281},
  year={2025},
  url     = {https://arxiv.org/abs/2505.23281}
}

@article{octothinker2025,
  title={{OctoThinker}: Mid-training Incentivizes Reinforcement Learning Scaling},
  author={Wang, Zengzhi and Zhou, Fan and Li, Xuefeng and Liu, Pengfei},
  journal={arXiv preprint arXiv:2506.20512},
  year={2025},
  url     = {https://arxiv.org/abs/2506.20512}
}

@article{song2026survey,
  title   = {A Survey of {On-Policy} Distillation for Large Language Models},
  author  = {Song, Mingyang and Zheng, Mao},
  journal = {arXiv preprint arXiv:2604.00626},
  year    = {2026},
  url     = {https://arxiv.org/abs/2604.00626}
}

@article{zhang2026formula,
  title   = {A {Formula-Driven} Survey and Research Agenda for {On-Policy} Distillation},
  author  = {Zhang, Bowen},
  journal = {arXiv preprint arXiv:2606.22793},
  year    = {2026},
  url     = {https://arxiv.org/abs/2606.22793}
}

@article{cui2026overview,
  title   = {A Brief Overview: {On-Policy} {Self-Distillation} In Large Language Models},
  author  = {Cui, Fangming and Li, Sunan and Li, Jiahong},
  journal = {arXiv preprint arXiv:2605.18141},
  year    = {2026},
  url     = {https://arxiv.org/abs/2605.18141}
}

@article{li2026localizing,
  title   = {Localizing Credit at the Divergence: Path-Conditioned Self-Distillation for {LLM} Reasoning},
  author  = {Li, Yu and Hong, Shu and Lan, Tian},
  journal = {arXiv preprint arXiv:2606.15576},
  year    = {2026},
  url     = {https://arxiv.org/abs/2606.15576}
}

\end{document}